\documentclass{ijac}
\usepackage{multicol}
\usepackage{subfigure}
\usepackage{amsmath}
\usepackage{amssymb}
\usepackage{amsfonts}
\usepackage{graphicx}
\usepackage{url}
\usepackage{ccaption}
\usepackage{booktabs}

\usepackage{times}
\usepackage{helvet}
\usepackage{courier}
\usepackage{graphicx}
\usepackage{amssymb,amsfonts,amsmath,dsfont,amsthm}
\usepackage[utf8]{inputenc} 
\usepackage[T1]{fontenc}    
\usepackage{url}            
\usepackage{booktabs}       
\usepackage{nicefrac}       
\usepackage{microtype}      
\usepackage[svgnames]{xcolor}
\usepackage{xcolor,colortbl} 
\providecommand{\scal}[2]{\left\langle{#1},{#2}\right\rangle}    

\newcommand{\C}{\mathcal{C}}

\renewcommand{\O}{\mathcal{O}}

\DeclareMathOperator{\R}{\mathbb{R}}

\providecommand{\scal}[2]{\left\langle{#1},{#2}\right\rangle}
\newcommand{\be}{\begin{equation}}
\newcommand{\ee}{\end{equation}}
\newcommand{\bt}{\begin{theorem}}
\newcommand{\et}{\end{theorem}}
\newcommand{\bd}{\begin{definition}}
\newcommand{\ed}{\end{definition}}
\newcommand{\br}{\begin{remark}}
\newcommand{\er}{\end{remark}}
\newcommand{\x}{\mathbf{x}}

\renewcommand{\k}{\mathbf{k}}
\newcommand{\XX}{\mathbb{X}}
\newtheorem{theorem}{Theorem}
\newtheorem{definition}{Definition}

\newtheorem{proposition}{Proposition}
\newtheorem{remark}{Remark}

\newtheorem{corollary}{Corollary}
\newtheorem{prob}{Problem}

\newenvironment{Proof}{\noindent{\sc Proof.}}{\qed}

\firstheadname{} 
\firstfootname{} 
\headevenname{} 
\headoddname{}

\renewcommand{\baselinestretch}{1}
\newcommand*{\titleAT}{\begingroup
  \newlength{\drop}
  \drop=0.05\textheight
  \begin{center}
  \includegraphics[scale=0.4]{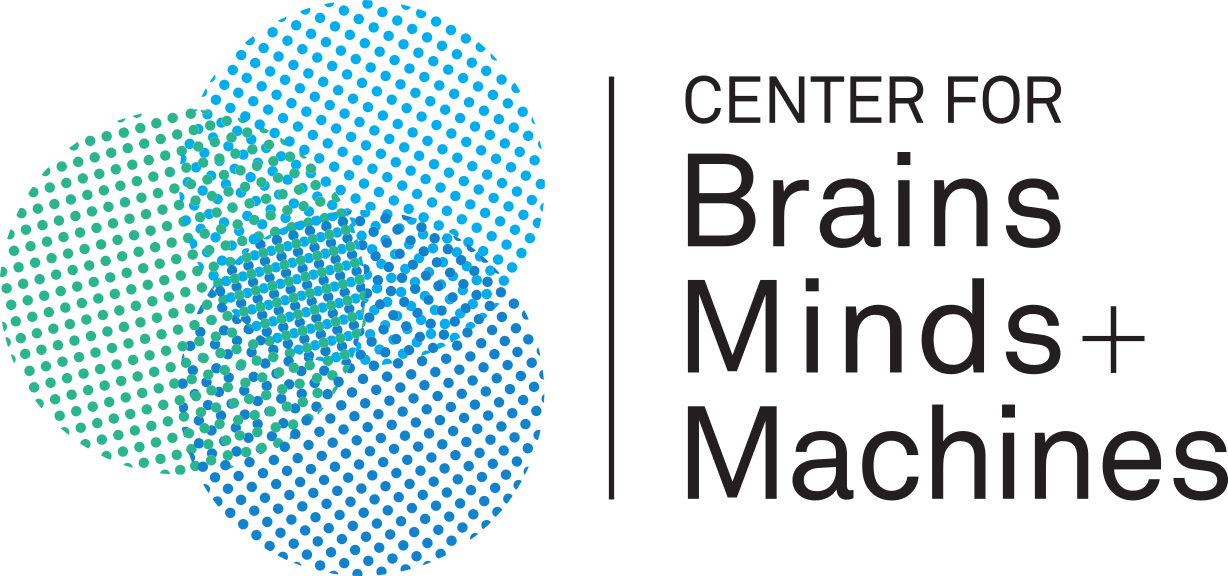} 
  \end{center} 
  \vspace{2pt}\vspace{-\baselineskip}

  \vspace{\drop}
  \textbf{\large{CBMM Memo No. \memonumber}}   \hfill    \textbf{\large{\memodate}} 

  \vspace{\drop}
  \begin{center}
    \textbf{\huge{\memotitle}}\\
    \vspace{0.4\drop}
    \textbf{\Large{by}}\\
    \vspace{0.4\drop}
    \large{\memoauthors}
  \end{center}
  \vspace{\drop}
  \textbf{\large{\noindent Abstract}:} {\memoabstract}

\vspace{\fill}
  \rule{\textwidth}{0.4pt}\par

  \begin{minipage}{.15\linewidth}
    \includegraphics[scale=0.1]{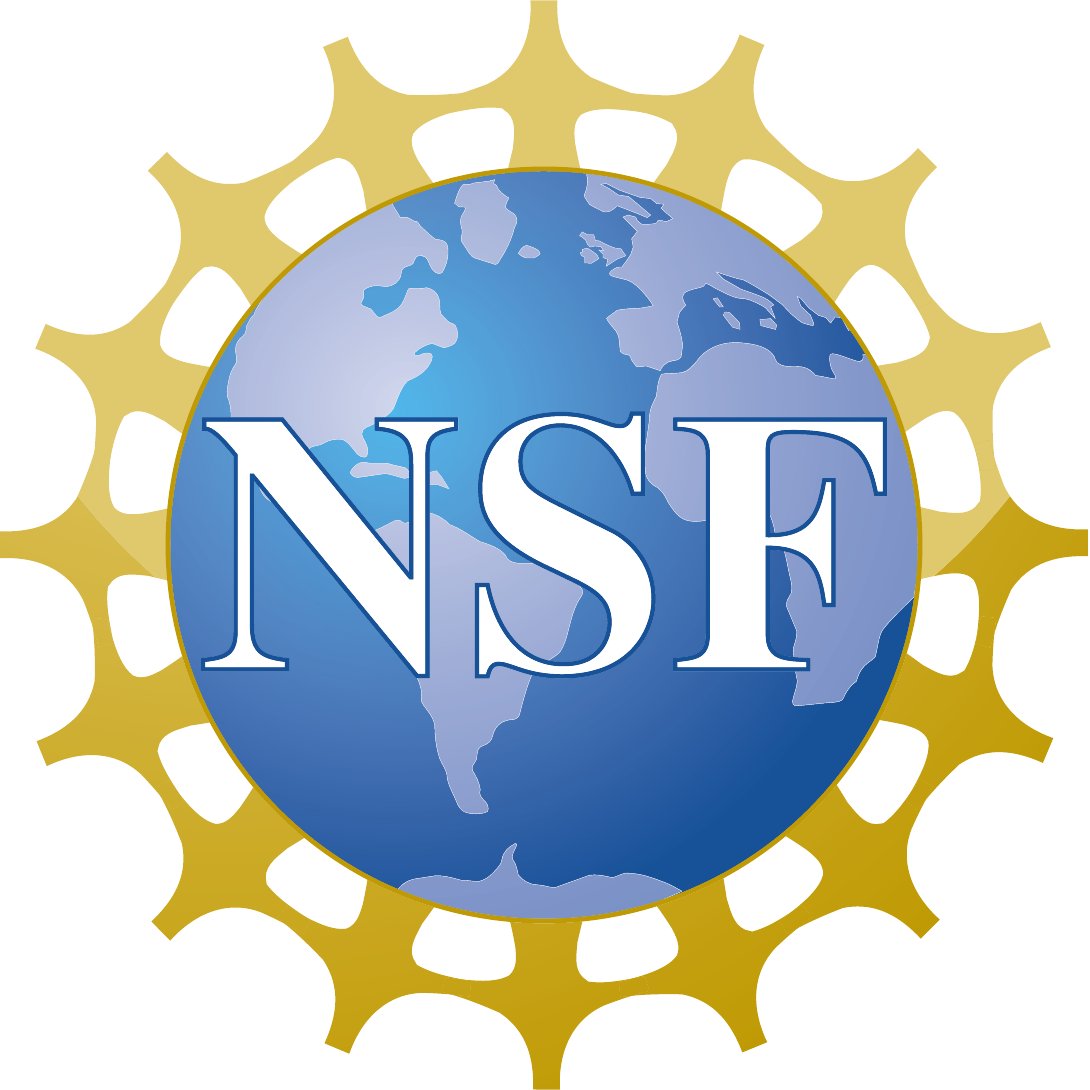}
  \end{minipage}
  \begin{minipage}{.84\linewidth}
    \textbf{\large{This work was supported by the Center for Brains, Minds and Machines (CBMM), funded by NSF STC award  CCF - 1231216.
        H.M. is supported in part by ARO Grant W911NF-15-1-0385.}}
  \end{minipage}
  \endgroup}

\def\RR{{\mathbb R}}

\def\PPI{{{\rm I}\kern-1pt\Pi}}

\def\b #1;{{\bf #1}}
\def\x{{\bf x}}
\def\k{{\bf k}}

\def\e{\epsilon}

\def\O{{\cal O}}

\def\C{{\mathcal C}}

\def\be{\begin{equation}}
\def\ee{\end{equation}}
\def\bea{\begin{eqnarray}}
\def\eea{\end{eqnarray}}
\def\eref#1{(\ref{#1})}

\def\donchitre#1#2{\vskip 6.5cm\noindent
\parbox[t]{1in}{\special{eps:#1.eps x=6.5cm y=5.5cm}}
\hbox to 7cm{}\parbox[t]{0.0cm}{\special{eps:#2.eps x=6.5cm y=5.5cm}}}

\def\XX{{\mathbb X}}

\begin{document}

\setcounter{page}{1}
\pagenumbering{arabic}
\onecolumn 
\def\memonumber{058}
\def\memodate{\today}
\def\memotitle{Why and When Can Deep – but Not Shallow – Networks Avoid the Curse of Dimensionality: a Review}
\def\memoauthors{ Tomaso Poggio$^{1}$ \qquad Hrushikesh Mhaskar$^{2}$ \qquad Lorenzo
Rosasco$^{1}$  \qquad Brando Miranda$^1$  \qquad Qianli Liao$^{1}$ \\
\vskip 0.3in
$^1$Center for Brains, Minds, and Machines, McGovern Institute for Brain Research, \\
   Massachusetts Institute of Technology, Cambridge, MA, 02139. 
\\
$^2$Department of Mathematics, California Institute of Technology, Pasadena, CA 91125; \\
Institute of Mathematical Sciences, Claremont Graduate University, Claremont, CA 91711 
}

\normalsize \def\memoabstract{The paper characterizes classes of
  functions for which deep learning can be exponentially better than
  shallow learning. Deep convolutional networks are a special case of
  these conditions, though weight sharing is not the main reason for
  their exponential advantage. }

\titleAT

\newpage

\title{Why and When Can Deep – but Not Shallow – Networks Avoid the Curse of Dimensionality: a Review}

\author{
Tomaso Poggio$^{1}$ \qquad Hrushikesh Mhaskar$^{2}$ \qquad Lorenzo
Rosasco$^{1}$  \qquad Brando Miranda$^1$  \qquad Qianli Liao$^{1}$ }

\address{
$^1$Center for Brains, Minds, and Machines, McGovern Institute for Brain Research,
   Massachusetts Institute of Technology, Cambridge, MA, 02139. 
\\
$^2$Department of Mathematics, California Institute of Technology, Pasadena, CA 91125;
Institute of Mathematical Sciences, Claremont Graduate University, Claremont, CA 91711 

}

\abstract{The paper characterizes classes of functions for which deep learning
  can be exponentially better than shallow learning. Deep
  convolutional networks are a special case of these conditions,
  though weight sharing is not the main reason for their exponential
  advantage. }


\maketitle

\pagestyle{ijacheadings}

\section{A theory of deep learning}

\subsection{Introduction}

There are at three main sets of theory questions about Deep
Neural Networks.  The first set of questions is about the power of the
architecture -- which classes of functions can it approximate and
learn well?  The second set of questions is about the learning
process: why is SGD (Stochastic Gradient Descent) so unreasonably
efficient, at least in appearance? The third, more important question
is about generalization.  Overparametrization may explain
why minima are easy to find during training but then why does overfitting
seems to be less of a problem than for classical shallow networks? Is
this because deep networks are very efficient algorithms for
Hierarchical Vector Quantization?

In this paper we focus especially on the first set of questions,
summarizing several theorems that have appeared online in
2015\cite{HierarchicalKernels2015,Hierarchical2015,poggio2015December},
and in 2016\cite{Mhaskaretal2016,MhaskarPoggio2016}). We
then describe additional results as well as a few conjectures and open
questions. The main message is that deep  networks have
the theoretical guarantee, which shallow networks do not have, that
they can avoid the {\it curse of dimensionality}  for an important class of problems,
corresponding to {\it compositional functions}, that is functions of
functions. An especially interesting subset of such compositional
functions are  {\it hierarchically local compositional
  functions} where all the constituent functions are local in the
sense of bounded small dimensionality. The deep networks that can
approximate them without the curse of dimensionality are of the deep
convolutional type (though weight sharing is not necessary). 

Implications of the theorems likely to be relevant in practice are:

\begin{enumerate}

\item Certain {\it deep convolutional architectures} have a theoretical
  guarantee that they can be {\it  much better} than one layer architectures such
  as kernel machines;
\item the problems for which certain deep networks are guaranteed to avoid
the curse of dimensionality (see for a nice review
  \cite{Donoho00high-dimensionaldata})  correspond to input-output
  mappings that are {\it compositional}. The most interesting set of
  problems consists of compositional functions composed of a hierarchy
  of constituent functions that are local: an example is
  $f(x_1, \cdots, x_8) = h_3(h_{21}(h_{11} (x_1, x_2), h_{12}(x_3,
  x_4)), \allowbreak h_{22}(h_{13}(x_5, x_6), h_{14}(x_7, x_8)))
  $. The compositional function $f$ requires only ``local''
  computations (here with just dimension $2$) in each of its
  constituent functions $h$;
\item the key aspect of convolutional networks that can give them an
  exponential advantage is {\it not weight sharing} but {\it locality} at each
  level of the hierarchy.

\end{enumerate}

\section{Previous theoretical work}

Deep Learning references start with Hinton's backpropagation and with
Lecun's convolutional networks (see for a nice review
\cite{LeCunBengioHinton2015}). Of course, multilayer convolutional
networks have been around at least as far back as the optical
processing era of the 70s. The Neocognitron\cite{fukushima:1980} was a
convolutional neural network that was trained to recognize characters.
The property of {\it compositionality} was a main motivation for
hierarchical models of visual cortex such as HMAX which can be
regarded as a pyramid of AND and OR layers\cite{Riesenhuber1999}, that
is a sequence of conjunctions and disjunctions.  Several papers in the
'80s focused on the approximation power and learning properties of
one-hidden layer networks (called shallow networks here). Very little
appeared on multilayer networks, (but see \cite{mhaskar1993approx,
  chui1994neural, chui1996}), mainly because one hidden layer nets
performed empirically as well as deeper networks. On the theory side,
a review by Pinkus in 1999\cite{Pinkus1999} concludes that ``...there
seems to be reason to conjecture that the two hidden layer model may
be significantly more promising than the single hidden layer
model...''.  A version of the questions about the importance of
hierarchies was asked in \cite{poggio03mathematics} as follows:
``\textit{A comparison with real brains offers another, and probably
  related, challenge to learning theory. The ``learning algorithms''
  we have described in this paper correspond to one-layer
  architectures. Are hierarchical architectures with more layers
  justifiable in terms of learning theory? It seems that the learning
  theory of the type we have outlined does not offer any general
  argument in favor of hierarchical learning machines for regression
  or classification.  This is somewhat of a puzzle since the
  organization of cortex -- for instance visual cortex -- is strongly
  hierarchical.  At the same time, hierarchical learning systems show
  superior performance in several engineering applications.}''
Because of the great empirical success of deep learning over the last
three years, several papers addressing the question of why hierarchies
have appeared.  Sum-Product networks, which are equivalent to
polynomial networks (see
\cite{Moore1988,DBLP:journals/corr/abs-1304-7045}), are a simple case
of a hierarchy that was analyzed\cite{DBLP:conf/nips/DelalleauB11} but
did not provide particularly useful insights.  Montufar and
Bengio\cite{MontufarBengio2014} showed that the number of linear
regions that can be synthesized by a deep network with ReLU
nonlinearities is much larger than by a shallow network. The meaning
of this result in terms of approximation theory and of our results is
at the moment an open question\footnote{We conjecture that the result
  may be similar to other examples in section \ref{lowerbounds}. It
  says that among the class of functions that are piecewise linear,
  there exist functions that can be synthesized by deep networks with
  a certain number of units but
  require a much large number of units to be synthesized by shallow
  networks}.  Relevant to the present review is the work on
hierarchical quadratic
networks\cite{DBLP:journals/corr/abs-1304-7045}, together with
function approximation results\cite{mhaskar1993neural,Pinkus1999}.
Also relevant is the conjecture by Shashua (see \cite{CohenS15}) on a
connection between deep learning networks and the hierarchical Tucker
representations of tensors. In fact, our theorems describe formally
the class of functions for which the conjecture holds.  This paper
describes and extends results presented
in\cite{Anselmi2014,anselmi2015theoretical,poggioetal2015} and
in\cite{LiaoPoggio2016, Mhaskaretal2016} which derive new upper bounds
for the approximation by deep networks of certain important classes of
functions which avoid the curse of dimensionality. The upper bound for
the approximation by shallow networks of general functions was well
known to be exponential. It seems natural to assume that, since there
is no general way for shallow networks to exploit a compositional
prior, lower bounds for the approximation by shallow networks of
compositional functions should also be exponential. In fact, examples
of specific functions that cannot be represented efficiently by
shallow networks have been given very recently by Telgarsky
\cite{Telgarsky2015} and by Shamir \cite{SafranShamir2016}. We provide
in theorem \ref{lower-bound} another example of a class of
compositional functions for which there is a gap between shallow and
deep networks.

\section{Function approximation by deep networks}\label{degapproxsect}

In this section, we state theorems about the approximation properties of 
shallow and deep networks. 

\subsection{Degree of approximation}\label{approxsect}

The general paradigm is as follows. We are interested in determining
how complex a network ought to be to {\it theoretically guarantee}
approximation of an unknown target function $f$ up to a given accuracy
$\epsilon>0$.  To measure the accuracy, we need a norm $\|\cdot\|$ on
some normed linear space $\mathbb{X}$. As we will see the norm used in
the results of this paper is the $sup$ norm in keeping with the
standard choice in approximation theory. Notice, however, that from
the point of view of machine learning, the relevant norm is the $L_2$
norm. In this sense, several of our results are stronger than
needed. On the other hand, our main results on compositionality
require the sup norm in order to be independent from the unknown
distribution of the inputa data. This is important  for  machine
learning.

\begin{figure*}
\centering
\includegraphics[width=0.9\textwidth]{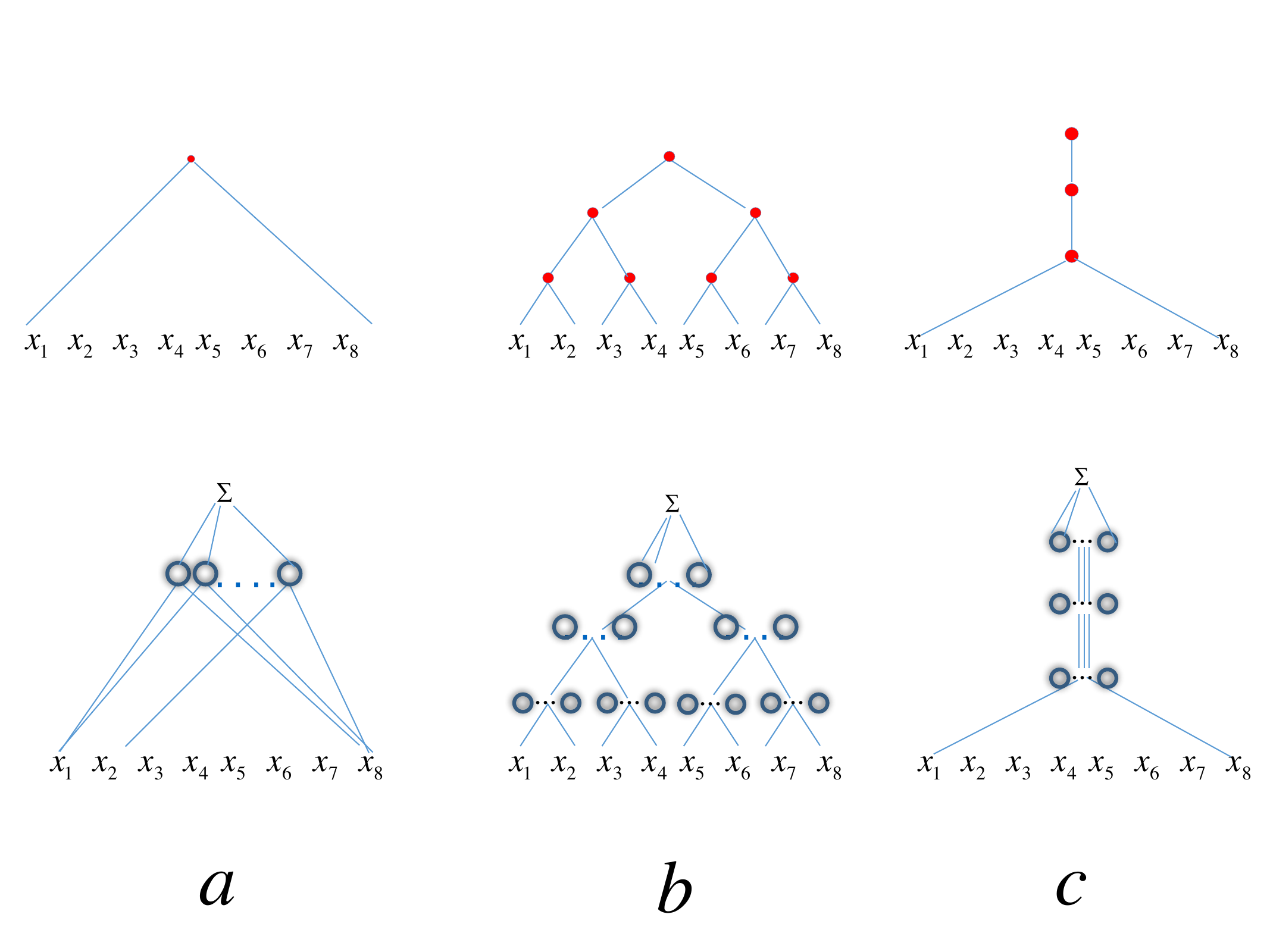}
\caption{The top graphs are associated to {\it functions}; each of the
  bottom diagrams depicts the ideal {\it network} approximating the
  function above. In a) a shallow universal network in 8 variables and
  $N$ units approximates a generic function of $8$ variables
  $f(x_1, \cdots, x_8)$. Inset b) shows a binary tree hierarchical
  network at the bottom in $n=8$ variables, which approximates well
  functions of the form
  $f(x_1, \cdots, x_8) = h_3(h_{21}(h_{11} (x_1, x_2), h_{12}(x_3,
  x_4)), \allowbreak h_{22}(h_{13}(x_5, x_6), h_{14}(x_7, x_8))) $ as
  represented by the binary graph above.  In the approximating network
  each of the $n-1$ nodes in the graph of the function corresponds to
  a set of $Q =\frac{N}{n-1}$ ReLU units computing the ridge function
  $\sum_{i=1}^Q a_i(\scal{\mathbf{v}_i}{\mathbf{x}}+t_i)_+$, with
  $\mathbf{v}_i, \mathbf{x} \in \R^2$, $a_i, t_i\in\R$. Each term in
  the ridge function corresponds to a unit in the node (this is
  somewhat different from todays deep networks, but equivalent to
  them, see text and note in \ref{Notes}). In a binary tree with $n$
  inputs, there are $log_2 n$ levels and a total of $n-1$
  nodes. Similar to the shallow network, a hierarchical network is
  universal, that is, it can approximate any continuous function; the
  text proves that it can approximate a compositional functions
  exponentially better than a shallow network.  No invariance -- that
  is weight sharing -- is assumed here. Notice that the key property
  that makes convolutional deep nets exponentially better than shallow
  for compositional functions is the locality of the constituent
  functions -- that is their low dimensionality. Weight sharing
  corresponds to all constituent functions at one level to be the same
  ($h_{11} = h_{12}$ etc.).  Inset c) shows a different mechanism that
  can be exploited by the deep network at the bottom to reduce the
  curse of dimensionality in the compositional function at the top:
  leveraging different degrees of smoothness of the constituent
  functions, see Theorem \ref{deepnetapprox} in the text. Notice that
  in c) the input dimensionality must be $\ge 2$ in order for deep
  nets to have an advantage over shallow nets. The simplest examples
  of functions to be considered for a), b) and c) are polynomials with
  a structure corresponding to the graph at the top.}
\label{example_3_functions}
\end{figure*}

Let $V_N$ be the be set of all networks of a given kind with
complexity $N$ which we take here to be the total number of units in
the network (e.g., all shallow networks with $N$ units in the hidden
layer). It is assumed that the class of networks with a higher
complexity include those with a lower complexity; i.e.,
$V_N\subseteq V_{N+1}$. The \textit{degree of approximation} is
defined by
\begin{equation}
\label{degapproxdef}
\mathsf{dist}(f, V_N)=\inf_{P\in V_N}\|f-P\|.
\end{equation}
For example, if $\mathsf{dist}(f, V_N)=\O(N^{-\gamma})$ for some
$\gamma>0$, then a network with complexity
$N=\O(\epsilon^{-\frac{1}{\gamma}})$ will be sufficient to guarantee an
approximation with accuracy at least $\epsilon$. Since $f$ is unknown, in order to obtain
theoretically proved upper bounds, we need to make some assumptions on
the class of functions from which the unknown target function is
chosen. This a priori information is codified by the statement that
$f\in W$ for some subspace $W\subseteq \mathbb{X}$. This subspace is
usually a smoothness class characterized by a smoothness parameter
$m$. Here it will be generalized to a smoothness and compositional
class, characterized by the parameters $m$ and $d$ ($d=2$ in the
example of Figure \ref{example_3_functions}; in general is the size of
the kernel in a convolutional network).

\subsection{Shallow and deep networks}
\label{subprevious}

This section characterizes conditions under which deep networks are
``better'' than shallow network in approximating functions.  Thus we
compare shallow (one-hidden layer) networks with deep networks as
shown in Figure \ref{example_3_functions}.  Both types of networks use
the same small set of operations -- dot products, linear combinations,
a fixed nonlinear function of one variable, possibly convolution and
pooling. Each node in the networks we consider usually corresponds to
a node in the graph of the function to be approximated, as shown in
the Figure. In particular each node in the network contains a certain
number of units. A unit is a neuron which computes

\begin{equation}
    (\scal{x}{w}+b)_+,
\label{noconv}
\end{equation}

\noindent where $w$ is the vector of weights on the vector input
$x$. Both $t$ and the real number $b$ are parameters tuned by
learning. We assume here that each node in the networks computes the
linear combination of $r$ such units

\begin{equation}
\sum_{i=1}^r c_i (\scal{x}{t_i}+b_i)_+
\label{ridge}
\end{equation}

Notice that for our main example of a deep network corresponding to a
binary tree graph, the resulting architecture is an idealized version
of the plethora of deep convolutional neural networks described in the
literature. In particular, it has only one output at the top unlike
most of the deep architectures with many channels and many top-level
outputs. Correspondingly, each node computes a single value instead of
multiple channels, using the combination of several units (see
Equation \ref{ridge}). Our approach and basic results apply rather
directly to more complex networks (see third note in section \ref{Notes}). A
careful analysis and comparison with simulations will be described in
future work.
\noindent

The logic of our theorems is as follows.

\begin{itemize}

\item {\it Both shallow (a) and deep (b) networks are universal}, that
  is they can approximate arbitrarily well any continuous function of
  $n$ variables on a compact domain. The result for shallow networks
  is classical. Since shallow networks can be viewed as a special case
  of deep networks, it clear that for any continuous function of $n$
  variables, there exists also a deep network that approximates the
  function arbitrarily well on a compact domain.

\item We consider a special class of functions of $n$ variables on a
  compact domain that are a hierarchical compositions of local functions
  such as

\begin{eqnarray}
  \lefteqn{f(x_1,
  \cdots, x_8) = h_3(h_{21}(h_{11} (x_1, x_2),  h_{12}(x_3, x_4)),}\nonumber\\
 && \qquad h_{22}(h_{13}(x_5, x_6),  h_{14}(x_7, x_8)))
 \label{l-variables}
\end{eqnarray}

\noindent The structure of the function in equation  \ref{l-variables}
is represented by a graph of the binary tree type. This is the simplest example of
compositional functions, reflecting dimensionality $d=2$ for the
constituent functions $h$. In general, $d$ is arbitrary but fixed and
independent of the dimensionality $n$ of the compositional function
$f$. In our results we will often think of $n$ increasing while $d$ is
fixed. In section \ref{EffectiveDimensionality} we will consider the
more general compositional case.

\item The approximation of functions with a {\it compositional
    structure} -- can be achieved with the same degree of accuracy by
  deep and shallow networks but that the number of parameters are much
  smaller for the deep networks than for the shallow network with
  equivalent approximation accuracy. It is intuitive that a
  hierarchical network matching the structure of a compositional
  function should be ``better'' at approximating it than a generic
  shallow network but universality of shallow networks asks for
  non-obvious characterization of ``better''.  Our result makes clear
  that the intuition is indeed correct.
\end{itemize}

In the perspetive of machine learning, we assume that the shallow
networks do not have any structural information on the function to be
learned (here its compositional structure), because they cannot
represent it directly and cannot exploit the advantage of a smaller
number of parameters. In any case, in the context of approximation
theory, we will exhibit and cite lower bounds of approximation by
shallow networks for the class of compositional functions. Deep
networks with standard architectures on the other hand {\it do
  represent} compositionality in their architecture and can be adapted
to the details of such prior information.

We approximate functions of $n$ variables of the form of Equation
(\ref{l-variables}) with networks in which the activation nonlinearity
is a smoothed version of the so called ReLU, originally called {\it
  ramp} by Breiman and given by $\sigma (x) = x_+ = max(0, x)$ .  The
architecture of the deep networks reflects Equation
(\ref{l-variables}) with each node $h_i$ being a ridge function,
comprising one or more neurons.

Let $I^n=[-1,1]^n$, $\XX=C(I^n)$ be the space of all continuous
functions on $I^n$, with $\|f\|=\max_{x\in I^n}|f(x)|$. 
Let
$\mathcal{S}_{N,n}$ denote the class of all shallow networks with $N$
units of the form
$$
x\mapsto\sum_{k=1}^N a_k\sigma(\scal{{w}_k}{x}+b_k),
$$
where ${w}_k\in\R^n$, $b_k, a_k\in\R$. The number of trainable
parameters here is $(n+2)N\sim n$. Let $m\ge 1$ be an integer, and
$W_m^n$ be the set of all functions of $n$ variables with continuous
partial derivatives of orders up to $m < \infty$ such that $\|f\|+\sum_{1\le
  |\k|_1\le m} \|D^\k f\| \le 1$, where $D^\k$ denotes the partial
derivative indicated by the multi-integer $\k\ge 1$, and $|\k|_1$ is
the sum of the components of $\k$. 

For the hierarchical binary tree network, the analogous spaces are
defined by considering the compact set $W_m^{n,2}$ to be the class of
all compositional functions $f$ of $n$ variables with a binary tree
architecture and constituent functions $h$ in $W_m ^2$.  We define the
corresponding class of deep networks $\mathcal{D}_{N,2}$ to be the set
of all deep networks with a binary tree architecture, where each of
the constituent nodes is in $\mathcal{S}_{M,2}$, where $N=|V|M$, $V$
being the set of non--leaf vertices of the tree. We note that in the
case when $n$ is an integer power of $2$, the total number of
parameters involved in a deep network in $\mathcal{D}_{N,2}$ -- that
is, weights and biases, is $4N$.

Two observations are critical to understand
the meaning of our results:

\begin{itemize}

\item compositional functions of $n$ variables are a subset of
  functions of $n$ variables, that is $W_m^n \supseteq W_m^{n,2}$.
  Deep networks can exploit in their architecture the special
  structure of compositional functions, whereas shallow networks are
  blind to it. Thus from the point of view of shallow networks,
  functions in $W_m^{n,2}$ are just functions in $W_m^n$; this is not
  the case for deep networks.

\item the deep network does not need to have exactly the same
  compositional architecture as the compositional function to be
  approximated. It is sufficient that the acyclic graph representing
  the structure of the function is {\it a subgraph of the graph
    representing the structure of the deep network}. The degree of
  approximation estimates depend on the graph associated with the
  network and are thus an upper bound on what could be achieved by a
  network exactly matched to the function architecture.

\end{itemize}

The following two theorems estimate the degree of approximation for
shallow and deep networks.

\subsection{Shallow networks}

The first theorem is about shallow networks.

\begin{theorem}
\label{optneurtheo}
Let $\sigma :\R\to \R$ be infinitely differentiable, and not a polynomial.  For $f\in W_m^n$ the complexity of shallow networks that
provide accuracy at least $\epsilon$ is 
\be 
N= \O(\epsilon^{-n/m})\,\, and\,\, 
is\,\, the\,\, best\,\, possible.  
\ee
\end{theorem}

\noindent\textit{Notes}
In \cite[Theorem~2.1]{optneur}, the theorem is stated under the
condition that $\sigma$ is infinitely differentiable, and there exists
$b\in\R$ such that $\sigma^{(k)}(b)\not=0$ for any integer $k\ge
0$. It is proved in \cite{corominas1954condiciones}
that the second condition is equivalent
to $\sigma$ not being a polynomial. The proof in \cite{optneur} relies
on the fact that under these conditions on $\sigma$, the algebraic
polynomials in $n$ variables of (total or coordinatewise) degree $<q$
are in the uniform closure of the span of $\O(q^n)$ functions of the
form $\x\mapsto\sigma(\scal{w}{x}+b)$ (see Appendix \ref{polapproach}).
The estimate itself is an upper bound on the degree of approximation
by such polynomials. Since it is based on the approximation of the
polynomial space contained in the ridge functions implemented by
shallow networks, one may ask whether it could be improved by using a
different approach. The answer relies on the concept of nonlinear
$n$--width of the compact set $W_m^n$ (cf.
\cite{devore1989optimal,Mhaskaretal2016}). The n-width results imply
that the estimate in Theorem (\ref{optneurtheo}) is {\it the best
  possible} among {\it all} reasonable \cite{devore1989optimal} methods of approximating
arbitrary functions in $W_m^n$.  $\square$
The estimate of Theorem \ref{optneurtheo} is the best possible if the only a priori
information we are allowed to assume is that the target function
belongs to $f\in W_m^n$. The exponential dependence on the
dimension $n$ of the number $\e^{-n/m}$ of parameters needed to
obtain an accuracy $\O(\epsilon)$ is known as the {\it curse of
  dimensionality}. Note that the constants involved in $\O$ in the theorems will depend upon
the norms of the derivatives of $f$ as well as $\sigma$.

A simple but useful corollary follows from the proof of Theorem
\ref{optneurtheo} about polynomials (which are a smaller space than
spaces of Sobolev functions). Let us denote with $P^n_k$ the linear space of
polynomials of degree at most $k$ in $n$ variables. Then

\begin{corollary}
\label{polynomial}
Let $\sigma :\R\to \R$ be infinitely differentiable, and not a
polynomial.  Every $f\in P^n_k$ can be realized with an arbitrary accuracy by shallow network
with $r$ units, $r=\binom{n+k}{k} \approx k^n $.
\end{corollary}

\subsection{Deep hierarchically local networks}

Our second and main theorem is about deep networks with smooth
activations and is recent (preliminary versions appeared in
\cite{poggio2015December,Hierarchical2015,Mhaskaretal2016}). We
formulate it in the binary tree case for simplicity but it extends
immediately to functions that are compositions of constituent
functions of a fixed number of variables $d$ instead than of $d=2$
variables as in the statement of the theorem (in convolutional
networks $d$ corresponds to the size of the kernel).

\begin{theorem}
\label{deeptheo}
For $f\in W_m^{n,2}$ consider a deep network with the same
compositonal architecture and with an activation function
$\sigma :\R\to \R$ which is infinitely differentiable, and
not a polynomial. The complexity of the
network to provide approximation with
accuracy at least $\epsilon$ is
\begin{equation}
N =\mathcal{O}((n-1)\epsilon^{-2/m}).
\label{deepnetapprox}
\end{equation}
\end{theorem}

\noindent\textit{Proof}
To prove Theorem~\ref{deeptheo}, we observe that each of the
constituent functions being in $W_m^2$, (\ref{optneurtheo}) applied
with $n=2$ implies that each of these functions can be approximated
from $\mathcal{S}_{N,2}$ up to accuracy $\epsilon=cN^{-m/2}$.  Our
assumption that $f\in W_m^{N,2}$ implies that each of these
constituent functions is Lipschitz continuous. Hence, it is easy to
deduce that, for example, if $P$, $P_1$, $P_2$ are approximations to
the constituent functions $h$, $h_1$, $h_2$, respectively within an
accuracy of $\epsilon$, then since $\|h-P\| \le \epsilon$,
$\|h_1-P_1\| \le \epsilon$ and $\|h_2-P_2\| \le \epsilon$, then
$\|h(h_1, h_2) - P(P_1, P_2)\| = \|h(h_1, h_2) - h(P_1, P_2) + h(P_1,
P_2) - P(P_1, P_2)\|\le \|h(h_1, h_2) - h(P_1, P_2)\| + \|h(P_1, P_2)
- P(P_1, P_2)\| \le c \epsilon$ by Minkowski inequality. Thus

$$
\|h(h_1,h_2)-P(P_1,P_2)\| \le c\epsilon,
$$
for some constant $c>0$ independent of the functions involved. This,
together with the fact that there are $(n-1)$ nodes,
leads to (\ref{deepnetapprox}). 
$\square$

Also in this case the proof provides the following corollary about the
subset $T^n_k$ of the space $P^n_k$ which consists of compositional
polynomials with a binary tree graph and constituent polynomial
functions of degree $k$ (in $2$ variables)

\begin{corollary}
\label{polynomial2}
Let $\sigma :\R\to \R$ be infinitely differentiable, and not a
polynomial. Let $n=2^l$. Then $f\in T^n_k$ can be realized by a deep
network with a binary tree graph and a total of $r$ units with
$r=(n-1) \binom{2+k}{2}\approx (n-1) k^2$.
\end{corollary}

It is important to emphasize that the assumptions on $\sigma$ in the
theorems are not satisfied by the ReLU function $x\mapsto x_+$, but
they are satisfied by smoothing the function in an arbitrarily small
interval around the origin. This suggests that the result of the
theorem should be valid also for the non-smooth ReLU. Section
\ref{non-smooth} provides formal results. Stronger results than the
theorems of this section (see \cite{MhaskarPoggio2016}) hold for
networks where each unit evaluates a Gaussian non--linearity; i.e.,
Gaussian networks of the form \be\label{gaussnetworkdef}
G(x)=\sum_{k=1}^N a_k\exp(-|x-w_k|^2),\qquad x\in\mathbb R^d \ee where
the approximation is on the entire Euclidean space.

In summary, when the only a priori assumption on the target function
is about the number of derivatives, then to {\it guarantee} an
accuracy of $\epsilon$, we need a shallow network with
$\O(\epsilon^{-n/m})$ trainable parameters. If we assume a
hierarchical structure on the target function as in
Theorem~\ref{deeptheo}, then the corresponding deep network yields a
guaranteed accuracy of $\epsilon$ with $\O(\epsilon^{-2/m})$ trainable
parameters. Note that Theorem~\ref{deeptheo} applies to all $f$ with a
compositional architecture given by a graph which correspond to, or is
a subgraph of, the graph associated with the deep network -- in this
case the graph corresponding to $W_m^{n,d}$. Theorem \ref{deeptheo}
leads naturally to the notion of {\it effective dimensionality} that we
formalize in the next section

\begin{definition}
\label{definition_effective_dimension}
  The {\it effective dimension} of a class $W$ of functions (for a
  given norm) is said to be $d$ if for every $\epsilon>0$, any
  function in $W$ can be recovered within an accuracy of $\epsilon$
  (as measured by the norm) using an appropriate network (either shallow or deep) with
  $\epsilon^{-d}$ parameters.
\end{definition}

Thus, the effective dimension for the class $W_m^n$ is $n/m$, that of $W_m^{n,2}$ is $2/m$.

  \section{General compositionality results: functions composed by a
    hierarchy of functions with bounded effective dimensionality}
\label{EffectiveDimensionality}

The main class of functions we considered in previous papers consists
of functions as in Figure \ref{example_3_functions} b that we called {\it compositional
  functions}. The term ``compositionality'' was used with the
meaning it has in language and vision, where higher level concepts are
composed of a small number of lower level ones, objects are composed
of parts, sentences are composed of words and words are composed of
syllables. Notice that this meaning of compositionality is narrower
than the mathematical meaning of composition of functions. The {\it
  compositional functions} we have described in previous papers may be
more precisely called {\it functions composed of hierarchically
  local functions}. 

\begin{figure}
\centering
\includegraphics[width=0.5\textwidth]{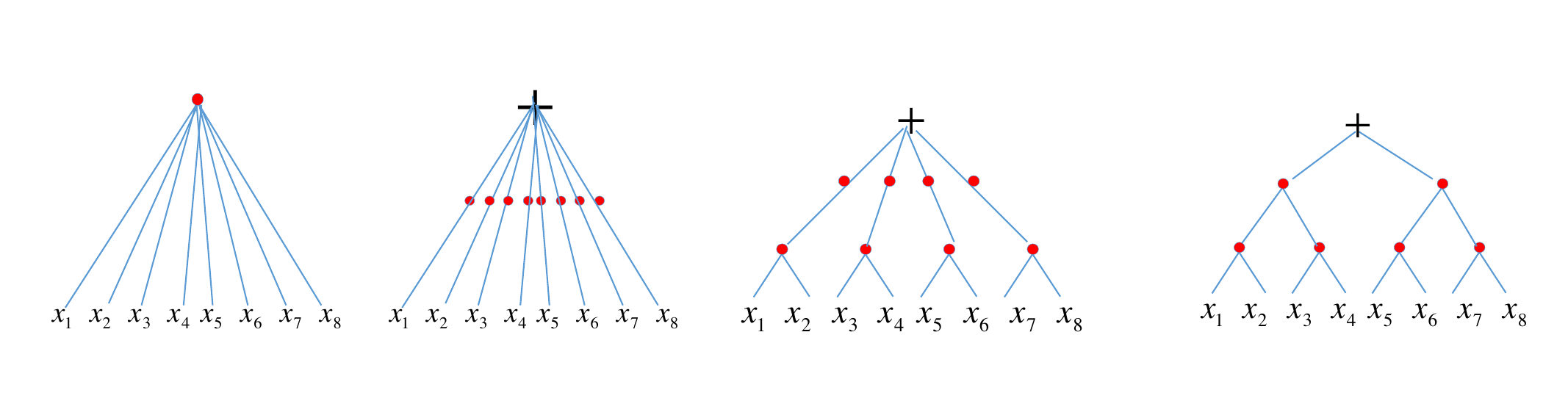}
\caption{ The figure shows the graphs of functions that may have small
  effective dimensionality, depending on the number of units per node
  required for good approximation.}
\label{LocalFunctions}
\end{figure}

Here we generalize formally our previous results to the broader class
of compositional functions (beyond the hierarchical locality of Figure
\ref{example_3_functions}b to Figure \ref{example_3_functions}c and
Figure \ref{LocalFunctions}) by restating formally a few comments of
previous papers. Let us begin with one of the previous examples. Consider

\begin{eqnarray*}
Q(x,y)&\!\!\!=\!\!\!&(Ax^2y^2+Bx^2y\\ 
         &                   &+Cxy^2+Dx^2+2Exy\\
         &                   &+  Fy^2+2Gx+2Hy+I)^{2^{10}}.
\end{eqnarray*}

Since $Q$ is nominally a polynomial of coordinatewise degree $2^{11}$,
  \cite[Lemma~3.2]{optneur} shows  that a shallow network with
$2^{11}+1$ units is able to approximate $Q$ arbitrarily well on
$I^2$. However, because of the hierarchical structure of $Q$,
  \cite[Lemma~3.2]{optneur} shows also that a hierarchical network
with $9$ units can approximate the quadratic expression, and $10$
further layers, each with $3$ units can approximate the successive
powers. Thus, a hierarchical network with $11$ layers and $39$ units
can approximate $Q$ arbitrarily well. We note that even if $Q$ is
nominally of degree $2^{11}$, each of the monomial coefficients in $Q$
is a function of only $9$ variables, $A,\cdots, I$.

A different example is 
\begin{equation}
Q(x,y)= |x^2 - y^2|.
\end{equation}
This is obviously a Lipschitz continuous function of $2$
variables. The effective dimension of this class is $2$, and hence, a
shallow network would require at least $c\epsilon^{-2}$ parameters to
approximate it within $\epsilon$. However, the effective dimension of
the class of univariate Lipschitz continuous functions is $1$.  Hence,
if we take into account the fact that $Q$ is a composition of a
polynomial of degree $2$ in $2$ variables and the univariate Lipschitz
continuous function $t\mapsto |t|$, then it is easy to see that the
same approximation can be achieved by using a two layered network with
$\O(\epsilon^{-1})$ parameters.

To formulate our most general result that includes the examples above
as well as the constraint of hierarchical locality, we first define formally a
compositional function in terms of a directed acyclic graph. Let
$\mathcal{G}$ be a directed acyclic graph (DAG), with the set of nodes
$V$. A $\mathcal{G}$--function is defined as follows. Each of the
source node obtains an input from $\R$. Each in-edge of every other
node represents an input real variable, and the node itself represents
a function of these input real variables, called a {\it constituent
  function}. The out-edges fan out the result of this evaluation.  We
assume that there is only one sink node, whose output is the
$\mathcal{G}$-function. Thus, ignoring the compositionality of this
function, it is a function of $n$ variables, where $n$ is the number
of source nodes in $\mathcal{G}$.

\begin{theorem}
\label{toptheo}
Let $\mathcal{G}$ be a DAG,$n$ be the number of source nodes, and for
each $v\in V$, let $d_v$ be the number of in-edges of $v$.  Let
$f: \R^n \mapsto \R$ be a compositional $\mathcal{G}$-function, where
each of the constitutent function is in $W_{m_v}^{d_v}$.  Consider
shallow and deep networks with infinitely smooth activation function
as in Theorem~\ref{optneurtheo}. Then deep networks -- with an
associated graph that corresponds to the graph of $f$ -- avoid the
curse of dimensionality in approximating $f$ for increasing $n$,
whereas shallow networks cannot directly avoid the curse. In
particular, the complexity of the best approximating shallow network
is exponential in $n$

\begin{equation}
N_s =\mathcal{O}(\epsilon^{-\frac{n} {m}}),
\label{shallow_approx}
\end{equation}
where $m=\min_{v\in V} m_v$,
\noindent while the complexity of the deep network is
\begin{equation}
N_d =\mathcal{O}(\sum_{v\in V} \epsilon^{- d_v/m_v}).
\label{deep_approx}
\end{equation}
\end{theorem}

Following definition \ref{definition_effective_dimension} we call
$d_v/m_v$ the {\it effective dimension} of function $v$. Then, deep
networks can avoid the curse of dimensionality if the constituent
functions of a compositional function have a small effective
dimension; i.e., have fixed, ``small'' dimensionality or fixed,
``small'' ``roughness. A different interpretation of
Theorem~\ref{toptheo} is the following.

\begin{proposition}
  If a family of functions $f: \R^n \mapsto \R$ of smoothness $m$ has
  an effective dimension $<n/m$, then the functions are compositional
  in a manner consistent with the estimates in Theorem~\ref{toptheo}.
\end{proposition}

Notice that the functions included in this theorem are functions that
are either local or the composition of simpler functions or
both. Figure \ref{LocalFunctions} shows some examples in addition to
the examples at the top of Figure Figure \ref{example_3_functions}.

As before, there is a simple corollary for polynomial functions:

\begin{corollary}
\label{polynomial3}
Let $\sigma :R \to R $ be infinitely differentiable, and not a
polynomial. With the set up as in Theorem~\ref{toptheo}, let $f$ be
DAG polynomial; i.e., a DAG function, each of whose constituent
functions is a polynomial of degree $k$. Then $f$ can be represented
by a deep network with $\mathcal{O}(|V_N|k^d)$ units, where $|V_N|$ is
the number of non-leaf vertices, and $d$ is the maximal indegree of
the nodes.
\end{corollary}

For example, if $\mathcal{G}$ is a full binary tree with $2^n$ leaves,
then the nominal degree of the $\mathcal{G}$ polynomial as in
Corollary~\ref{polynomial3} is $k^{k^n}$, and therefore requires a
shallow network with $\mathcal{O}(k^{2k^n})$ units, while a deep
network requires only $\mathcal{O}(nk^2)$ units.

Notice that polynomials in $S^n_k$ are {\it sparse} with a number of
terms which is not exponential in $n$, that is it is not $O(k^n)$ but
linear in $n$ (that is $O(n k)$) or at most polynomial in $n$.

\subsection{Approximation results  for shallow and deep  networks with
(non-smooth) ReLUs}
\label{non-smooth}
The results we described so far use smooth activation functions.  We
already mentioned why relaxing the smoothness assumption should not
change our results in a fundamental way.  While studies on the
properties of neural networks with smooth activation abound, the
results on non-smooth activation functions are much more sparse.  Here
we briefly recall some of them.

In the case of shallow networks, the condition of a smooth activation
function can be relaxed to prove density (see \cite{Pinkus1999},
Proposition 3.7):

\begin{proposition}
\label{optneurtheo2}
Let $\sigma = :\R\to \R$ be in $\C^0$, and not a polynomial. Then
shallow networks are dense in $\C^0$.
\end{proposition}

In particular, ridge functions using ReLUs of the form
$\sum_{i=1}^r c_i (\scal{{w}_i}{{x}}+b_i)_+$, with
${w}_i, {x} \in \R^n$, $c_i, b_i\in\R$ are dense in
$\C$. 

Networks with non-smooth activation functions are expected to do
relatively poorly in approximating smooth functions such as
polynomials in the sup norm. ``Good'' degree of approximation rates
(modulo a constant)  have been proved in the $L_2$ norm.
Define ${\cal B}$ the unit ball in $\R^n$. Call $C^m(B^n)$ the set of
all continuous functions with continuous derivative up to degree $m$
defined on the unit ball. We define the Sobolev space $W^m_p$ as the
completion of $C^m(B^n)$ with respect to the Sobolev norm $p$ (see for
details \cite{Pinkus1999} page 168). We define the space ${\cal
  B}^m_p= \{f \colon f \in W^m_p, \|f \|_{m,p} \leq 1 \}$ and the
approximation error $E(B^m_2; H; L_2)=\inf_{g \in H} \|f-g \|_{L_2}$. It is shown in
\cite[Corollary~6.10]{Pinkus1999} that

\begin{proposition}
\label{optneurtheo3}
For  $M^r \colon f(x)=
\sum_{i=1}^r c_i (\scal{{w}_i}{{x}}+b_i)_+$ it holds 
$E(B_2^m; M_r; L_2) \leq C r ^{-\frac{m}{n}}$ for $m=1, \cdots, \frac{n+3}{2}$.
\end{proposition}

These  approximation results with respect to the $L^2$ norm  cannot
be applied to derive bounds for compositional networks. Indeed, in the
latter case, as we remarked already, estimates in the uniform norm are
needed to control the propagation of the errors from one layer to the
next, see Theorem~\ref{deeptheo}.  Results in this direction are given
in \cite{Mhaskar2004}, and more recently in \cite{Bach2014} and
\cite{MhaskarPoggio2016} (see Theorem 3.1).  In particular, using a result in
\cite{Bach2014} and following the proof strategy of
Theorem~\ref{deeptheo} it is possible to derive the following results
on the approximation of Lipshitz continuous functions with deep and
shallow ReLU networks that mimics our Theorem \ref{deeptheo}:
\begin{theorem}
\label{bach}
Let $f$ be a $L$-Lipshitz continuous function of $n$ variables. 
Then, the complexity of  a network which is a linear combination of ReLU providing an
approximation with accuracy at least $\epsilon$ is 
$$
N_{s} =\mathcal{O}\left (\left (\frac{\epsilon}{L}\right)^{-n}\right),
$$
wheres that of a deep compositional architecture is
$$
N_{d} =\mathcal{O}\left (\left (n-1) (\frac{\epsilon}{L}\right) ^{-2}\right).
$$
\end{theorem}

Our general Theorem \ref{toptheo} can be extended in a similar way.
Theorem \ref{bach} is an example of how the analysis of smooth
activation functions can be adapted to ReLU.  Indeed, it shows how
deep compositional networks with standard ReLUs can avoid the curse of
dimensionality.  In the above results, the regularity of the function
class is quantified by the magnitude of Lipshitz constant.  Whether
the latter is the best notion of smoothness for ReLU based networks, and
if the above estimates can be improved, are interesting questions that
we defer to a future work. A result that is more intuitive and may
reflect what networks actually do is described in Appendix
\ref{Booleanization}. Though the construction described there 
provides approximation in the $L_2$ norm but not in the sup norm, this
is not a problem under any discretization of real number required for
computer simulations (see Appendix).

\begin{figure*}
\centering
\includegraphics[width=0.7\textwidth]{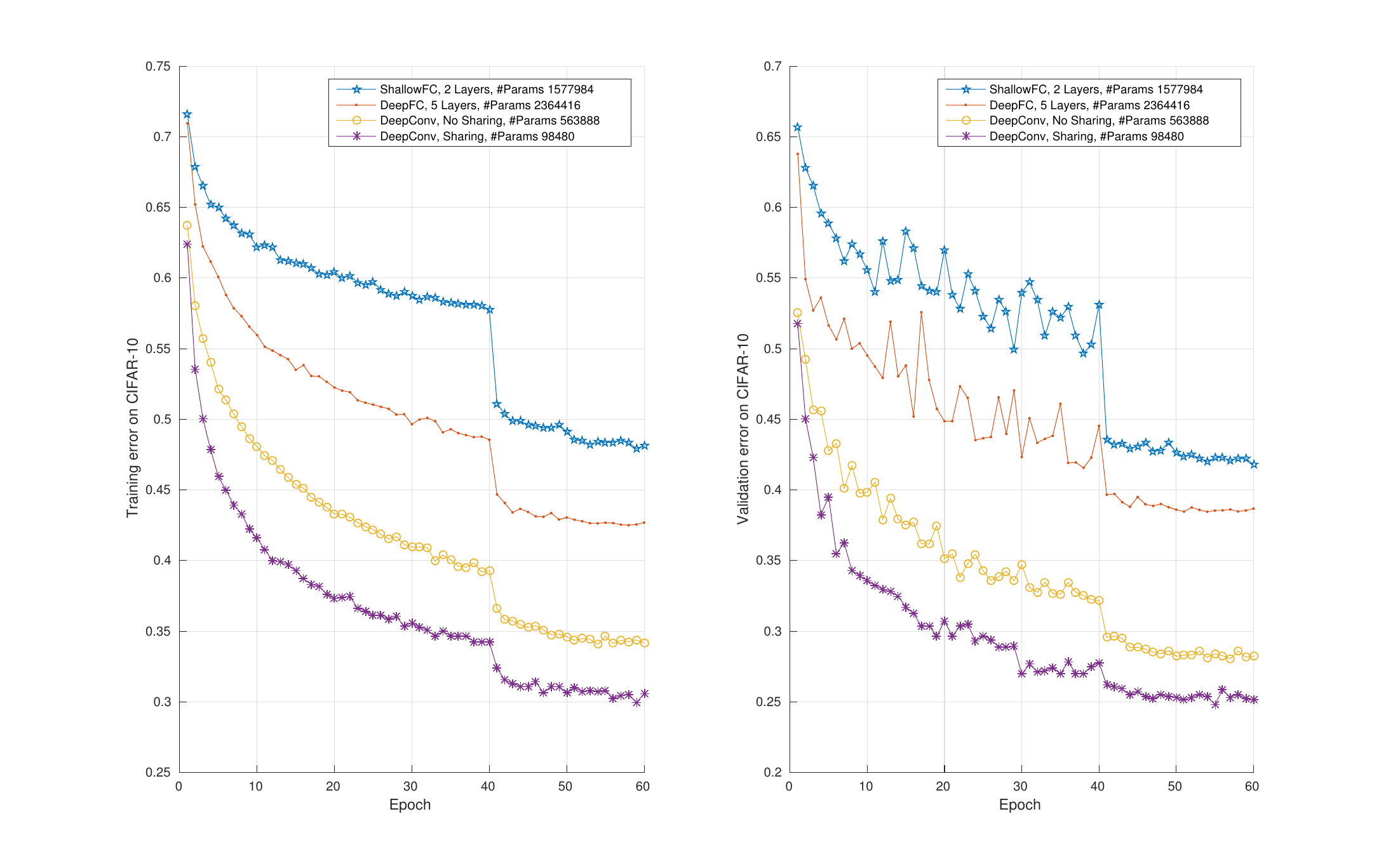}
\caption{We show that the main advantage of deep Convolutional
  Networks (ConvNets) comes from "hierarchical locality" instead of
  weight sharing. We train two 5-layer ConvNets with and without
  weight sharing on CIFAR-10. ConvNet without weight sharing has
  different filter parameters at each spatial location. There are 4
  convolutional layers (filter size 3x3, stride 2) in each network.
  The number of feature maps (i.e., channels) are 16, 32, 64 and 128
  respectively. There is an additional fully-connected layer as a
  classifier. The performances of a 2-layer and 5-layer
  fully-connected networks are also shown for comparison. Each hidden
  layer of the fully-connected network has 512 units. The models are
  all trained for 60 epochs with cross-entropy loss and standard shift
  and mirror flip data augmentation (during training). The training
  errors are higher than those of validation because of data
  augmentation. The learning rates are 0.1 for epoch 1 to 40, 0.01 for
  epoch 41 to 50 and 0.001 for rest epochs. The number of parameters
  for each model are indicated in the legends. Models with
  hierarchical locality significantly outperform shallow and
  hierarchical non-local networks.}
\label{conv_weight_sharing}
\end{figure*}

Figures \ref{Functions_Simulations}, \ref{Simulations},
  \ref{Simulations_relu2units}, \ref{conv_weight_sharing}
provide a sanity check and empirical support for our main results and
for the claims in the introduction.

\begin{figure}
\centering
\includegraphics[width=0.5\textwidth]{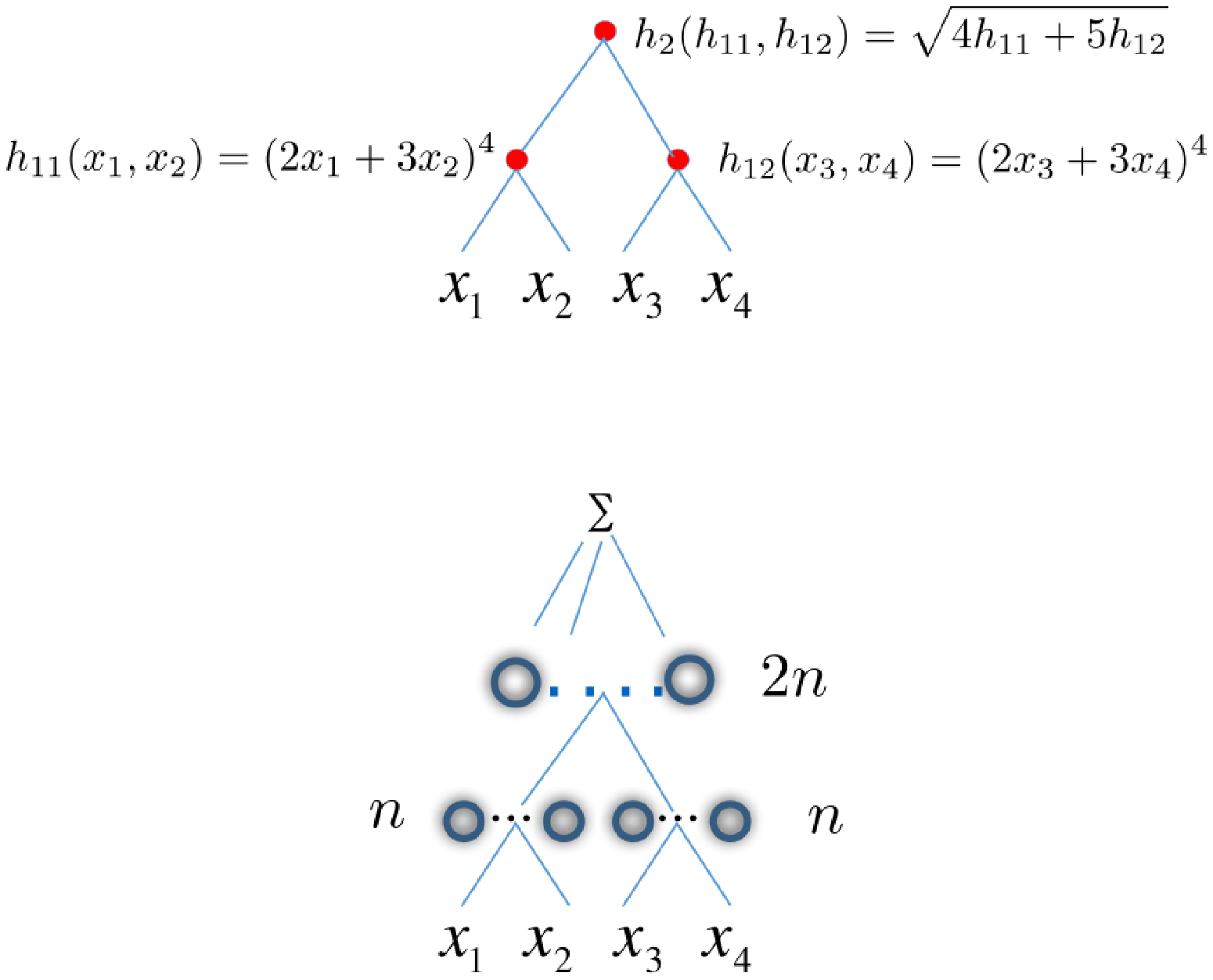}
\caption{ The figure shows on the top the graph of the function to be approximated,
  while the bottom part of the figure shows a deep neural network with
  the same graph structure. The left
  and right node inf the first layer has each $n$ units giving a total of
  $2n$ units in the first layer.  The second layer has a total of $2n$
  units.  The first layer has a convolution of size $n$ to mirror the
  structure of the function to be learned.  The compositional function
  we  approximate has the form
  $f(x_1, x_2, x_3, x_4) = h_2( h_{11}(x_1, x_2) , h_{12}(x_3, x_4) )
  $ with $h_{11}$, $h_{12}$ and $h_2$ as indicated in the figure.  }
\label{Functions_Simulations}
\end{figure}

\begin{figure*}
\centering
\includegraphics[width=1.0\textwidth]{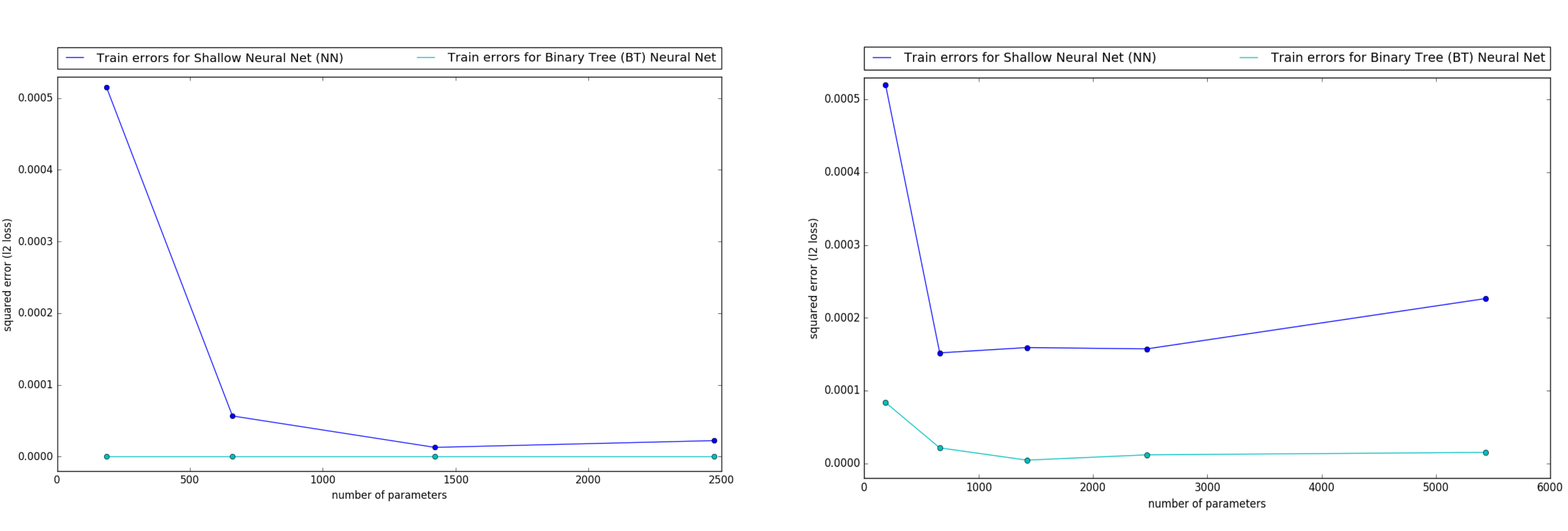}
\caption{An empirical comparison of shallow vs 2-layers binary tree
  networks in the approximation of compositional functions. The loss
  function is the standard mean square error (MSE). There are several
  units per node of the tree. In our setup here the network with an
  associated binary tree graph was set up so that each layer had the
  same number of units and shared parameters.  The number of units for
  the shallow and binary tree neural networks had the same number of
  parameters.  On the left the function is composed of a single ReLU
  per node and is approximated by a network using ReLU activations.
  On the right the compositional function is
  $f(x_1, x_2, x_3, x_4) = h_2( h_{11}(x_1, x_2) , h_{12}(x_3, x_4) )$
  and is approximated by a network with a smooth ReLU activation (also
  called softplus).  The functions $h_1$, $h_2$, $h_3$ are as
  described in Figure \ref{Functions_Simulations}.  In order to be
  close to the function approximation case, a large data set of 60K
  training examples was used for both training sets.  We used for SGD
  the Adam\cite{DBLP:journals/corr/KingmaB14} optimizer.  In order to
  get the best possible solution we ran 200 independent hyper
  parameter searches using random search \cite{randomhp} and reported
  the one with the lowest training error.  The hyper parameters search
  was over the step size, the decay rate, frequency of decay and the
  mini-batch size.  The exponential decay hyper parameters for Adam
  were fixed to the recommended values according to the original paper
  \cite{DBLP:journals/corr/KingmaB14}.  The implementations were based
  on TensorFlow \cite{tensorflow2015-whitepaper}.}
\label{Simulations} 
\end{figure*}

\begin{figure*}
\centering
\includegraphics[width=0.9\textwidth]{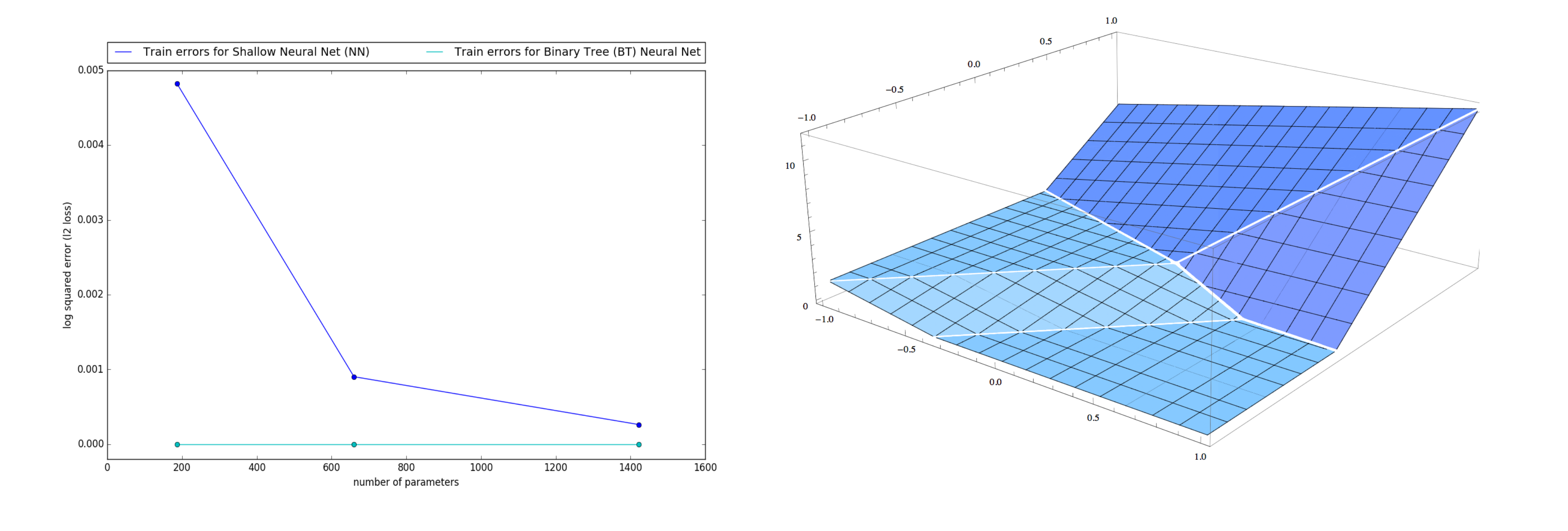}
\caption{Another comparison of shallow vs 2-layers binary tree
  networks in the learning of compositional functions. 
  The set up of the experiment was the same as in the one in figure \ref{Simulations} except that the
  compositional function had two ReLU units per node instead of only one. 
 The right part of the figure shows a cross section of the
  function  $f(x_1, x_2, 0.5, 0.25)$ in a bounded interval $x_1 \in
  [-1, 1], x_2 \in [-1, 1]$. The shape of the function is piecewise
  linear as it is always the case for ReLUs networks.}
\label{Simulations_relu2units}
\end{figure*}

\subsection{Lower bounds and gaps}
\label{lowerbounds} 

So far we have shown that there are deep networks -- for instance of
the convolutional type -- that can {\it avoid the curse of
  dimensionality} if the functions they are learning are {\it blessed
  with compositionality}.  There are no similar guarantee for shallow
networks: for shallow networks approximating generic continuous
functions the lower and the upper bound are both exponential\cite
{Pinkus1999}.  From the point of view of machine learning, it is
obvious that shallow networks, unlike deep ones, cannot exploit in
their architecture the reduced number of parameters associated with
priors corresponding to compositional functions. In past papers we
listed a few examples, some of which are also valid lower bounds from
the point of view of approximation theory:
\begin{itemize}

\item The polynomial considered earlier
\begin{eqnarray*}
Q(x_1,x_2,x_3,x_4)=(Q_1(Q_2(x_1,x_2), Q_3(x_3,x_4)))^{1024},
\end{eqnarray*}
\noindent can be approximated by deep networks with a smaller number
of parameters than shallow networks is based on polynomial
approximation of functions of the type $g(g(g()))$. Here, however, a
formal  proof of the impossibility of good approximation by
shallow networks is not available. For a lower bound we need at least  one
example of a compositional function which cannot be approximated by
shallow networks with a non-exponential degree of approximation.

\item Such an example, for which a proof of the lower bound exists
  since a few decades, consider
  a function which is a linear combination of $n$ tensor product
  Chui--Wang spline wavelets, where each wavelet is a tensor product
  cubic spline. It is shown in \cite{chui1994neural, chui1996} that is
  impossible to implement such a function using a shallow neural
  network with a sigmoidal activation function using $\O(n)$ neurons,
  but a deep network with the activation function $(x_+)^2$ can do
  so. In this case, as we mentioned, there is a formal proof of a gap
  between deep and shallow networks. Similarly, Eldan and Shamir
  \cite{EldanShamir2016} show other cases with separations that are
  exponential in the input dimension.

\item As we mentioned earlier, Telgarsky proves an exponential gap
  between certain functions produced by deep networks and their
  approximation by shallow networks.  The theorem \cite{Telgarsky2015}
  can be summarized as saying that {\it a certain family of
    classification problems with real-valued inputs cannot be
    approximated well by shallow networks with fewer than
    exponentially many nodes whereas a deep network achieves zero
    error}. This corresponds to high-frequency, sparse trigonometric
  polynomials in our case.  His upper bound can be proved directly
  from our main theorem by considering the real-valued polynomials
  $x_1x_2...x_d$ defined on the cube $(-1, 1)^d$ which is obviously a
  compositional function with a binary tree graph.

\item We exhibit here another example to illustrate a limitation of shallow networks in approximating a compositional function.

Let $n\ge 2$ be an integer, $B\subset \RR^n$ be the unit ball of
$\RR^n$. We consider the class $W$ of all compositional functions
$f=f_2\circ f_1$, where $f_1:\RR^n\to\RR$, and
$\sum_{|\k|\le 4}\|D^\k f_1\|_\infty \le 1$, $f_2: \RR\to\RR$ and
$\|D^4f_2\|_\infty\le 1$. We consider
$$
\Delta(\mathcal{A}_N):=\sup_{f\in W}\inf_{P\in \mathcal{A}_N}\|f-P\|_{\infty, B},
$$
where $\mathcal{A}_N$ is either the class $\mathcal{S}_N$ of all
shallow networks with $N$ units or $\mathcal{D}_N$ of deep networks
with two layers, the first with $n$ inputs, and the next with one
input. The both cases, the activation function is a $C^\infty$
function $\sigma :\RR\to\RR$ that is not a polynomial.
\begin{theorem}
\label{lower-bound}
There exist constants $c_1>0$ such that for $N\ge c_1$,
\be\label{shallowlowbd}
\Delta(\mathcal{S}_N) \ge \lceil 2^{-N/(n-1)}\rceil,
\ee
In contrast,  there exists $c_3>0$ such that
\be\label{deepupbd}
\Delta(\mathcal{D}_N) \le c_3N^{-4/n}.
\ee
The constants $c_1, c_2, c_3$ may depend upon $n$.
\end{theorem}

\begin{Proof}
  The estimate \eref{deepupbd} follows from the estimates already
  given for deep networks.    To prove
  \eref{shallowlowbd}, we use Lemma~3.2 in \cite{chui1996}.
  Let $\phi$ be a $C^\infty$ function supported on $[0,1]$, and we
  consider $f_N(x)=\phi(|4^Nx|^2)$. We may clearly choose $\phi$ so that 
  $\|f_N\|_\infty =1$. Then it is clear that each $f_N\in W$. Clearly, 
  \be\label{pf1eqn1}
  \Delta(\mathcal{S}_N) \ge  \inf_{P\in \mathcal{S}_N}
  \max_{x\in B}|f_N(x)-P(x)|.  
  \ee 
  We choose
  $P^*(x)=\sum_{k=1}^N \sigma(\scal{w_k^*}{x} +b_k^*)$ such that
  \be\label{pf1eqn2} 
  \inf_{P\in
    \mathcal{S}_N}\max_{x\in B}|f_N(x)-P(x)|\ge (1/2)
  \max_{x\in B}|f_N(x)-P^*(x)|.  
  \ee 
  Since $f_N$ is supported on
  $\{x\in\RR^n : |x|\le 4^{-N}\}$, we may imitate the proof of Lemma~3.2 in
  \cite{chui1996} with $g_k^*(t)=\sigma(t+b_k^*)$. Let $x_0\in B$ be such that (without loss of generality) $f_N(x_0)=\max_{x\in B}|f_N(x)|$, and $\mu_0$ be the Dirac measure supported at $x_0$. We group $\{w_k^*\}$ in $m=\lceil N/(n-1)\rceil$ disjoint groups of $n-1$ vectors each. For each group, we take vectors $\{v_\ell\}$ such that $v_\ell$ is orthogonal to the $w_k^*$'s in group $\ell$. The argument in the proof of 
Lemma~3.2 in
  \cite{chui1996} can be modified to get a measure $\mu$ with total variation $2^m$ such that
  $$
  \int_B f_N(x)d\mu(x)=\|f_N\|_\infty, \ \int_B g_k^*(x)d\mu(x)=0, \ k=1,\cdots,N.
 $$
 It is easy to deduce from here as in \cite{chui1996} using the duality principle that 
$$
  \max_{x\in B}|f_N(x)-P^*(x)|\ge c2^{-m}.
$$

Together with \eref{pf1eqn1} and \eref{pf1eqn2}, this implies \eref{shallowlowbd}.
\end{Proof}
\end{itemize}

So by now plenty of examples of lower bounds exist showing a gap between
shallow and deep networks. A particularly interesting case is the {\it
  product} function, that is the monomial
$f(x_1, \cdots, x_n) = x_1 x_2  \cdots x_n$ which
is, from our point of view, {\it the} prototypical compositional
functions. Keeping in mind the issue of lower bounds, the question
here has to do with the minimum integer $r(n)$ such that the function
$f$ is in the closure of the span of
$\sigma(\scal{w_k}{x} +b_k)$, with $k=1,\cdots, r(n)$, and $w_k$, $b_k$
ranging over their whole domains.  Such a result has been claimed for
the case of smooth ReLUs, using unusual group techniques and is
sketched in the Appendix of \cite{TegmarkLin2016}.


Notice, in support of the conjecture, that assuming that a shallow network
with (non-smooth) ReLUs has a lower bound of $r(q)=O(q)$ will lead to
an apparent contradiction with Hastad theorem (which is about
representation not approximation of Boolean functions) by restricting $x_i$ from
$x_i \in (-1,1)$ to $x_i \in \{-1,+1\}$. Hastad theorem
\cite{Hastad1987} establishes the inapproximability of the parity
function by shallow circuits of non-exponential size.

\subsection{Messy graphs and densely connected deep networks}

As mentioned already, the approximating deep network does not need to
exactly match the architecture of the compositional function as long
as the graph or tree associated with the function is contained in the
graph associated with the network.  This is of course good news: the
compositionality prior embedded in the architecture of the network
does not to reflect exactly the graph of a new  function to be
learned.  We have shown that for a given class of compositional
functions characterized by an associated graph there exist a deep
network that approximates such a function better than a shallow
network. The same network approximates well functions characterized by
subgraphs of the original class.

The proofs of our theorems show that linear combinations of
compositional functions are {\it universal} in the sense that they can
approximate any function and that deep networks with a number of units
that increases exponentially with layers can approximate any
function. Notice that deep compositional networks can interpolate if
they are overparametrized with respect to the data, even if the data
reflect a non-compositional function.

As an aside, note that the simplest compositional function -- addition --
is trivial in the sense that it offers no approximation advantage to
deep networks. The key function is multiplication which is for us the
prototypical compositional functions. As a consequence,  {\it polynomial functions are
  compositional} -- they are linear combinations of monomials which
are compositional. However, their compositional structure does not
confer any advantage in terms of approximation, because of the exponential
number of compositional terms.

As we mentioned earlier, networks corresponding to graphs that include
the graph of the function to be learned can exploit
compositionality. The relevant number of parameters to be optimized,
however, is the number of parameters $r$ in the network and not the
number of parameters $r*$ ($r* < r$) of the optimal deep network with
a graph exactly matched to the graph of the function to be learned. As
an aside, the price to be paid in using a non-optimal prior depend on
the learning algorithm. For instance, under sparsity constraints it
may be possible to pay a smaller price than $r$ (but higher than
$r^*$).

In this sense, some of the densely connected deep networks used in
practice -- which contain sparse graphs possibly relevant for the
function to be learned and which are still ``smaller'' than the
exponential number of units required to represent a generic function
of $n$ variables -- may be capable in some cases of exploiting an
underlying compositionality structure without paying an exhorbitant
price in terms of required complexity.

\section{Connections with the theory of Boolean functions}

The approach followed in our main theorems suggest the following
considerations (see Appendix \ref{Boolean Functions} for a brief
introduction).  The structure of a deep network is reflected in
polynomials that are best approximated by it -- for instance generic
polynomials or sparse polynomials (in the coefficients) in $d$
variables of order $k$. The tree structure of the nodes of a deep
network reflects the structure of a specific sparse
polynomial. Generic polynomial of degree $k$ in $d$ variables are
difficult to learn because the number of terms, trainable parameters
and associated VC-dimension are all exponential in $d$. On the other
hand, functions approximated well by sparse polynomials can be learned
efficiently by deep networks with a tree structure that matches the
polynomial.  We recall that in a similar way several properties of
certain Boolean functions can be ``read out'' from the terms of their
Fourier expansion corresponding to ``large'' coefficients, that is
from a polynomial that approximates well the function.

Classical results \cite{Hastad1987} about the depth-breadth tradeoff
in circuits design show that deep circuits are more efficient in
representing certain Boolean functions than shallow circuits. Hastad
proved that highly-variable functions (in the sense of having high
frequencies in their Fourier spectrum), in particular the parity
function cannot even be decently approximated by small constant depth
circuits (see also \cite{LinialMansour1993}). A closely related
result follow immediately from our main theorem since functions of
real variables of the form $x_1 x_2...x_d$ have the compositional form
of the binary tree (for $d$ even). Restricting the values of the
variables to $-1, +1$ yields an upper bound:

\begin{proposition}
The family of parity functions $x_1 x_2...x_d$ with $x_i \in \{-1,
+1\}$ and $i=1, \cdots, x_d$ can be represented with exponentially
fewer units by a deep than a shallow network.
\end{proposition}

Notice that Hastad's results on Boolean functions have been often quoted in support
of the claim that deep neural networks can represent functions that
shallow networks cannot. For instance Bengio and LeCun
\cite{BengioLecun2007} write {\it ``We claim that most functions that
  can be represented compactly by deep architectures cannot be
  represented by a compact shallow architecture''.''}.

Finally, we want to mention a few other observations on Boolean
functions that  shows an interesting connection with our approach. It is known
that within Boolean functions the $AC^0$ class of polynomial size
constant depth circuits is characterized by Fourier transforms where
most of the power spectrum is in the low order coefficients. Such
functions can be approximated well by a polynomial of low degree and
can be learned well by considering only such coefficients. There are 
two algorithms \cite{Mansour1994} that allow learning of certain
Boolean function classes:
\begin{enumerate}
\item the low order algorithm that approximates functions by
  considering their low order Fourier coefficients and
\item the sparse algorithm which learns a function by approximating
  its significant coefficients.
\end{enumerate}
Decision lists and decision trees can be learned by the first
algorithm. Functions with small $L_1$ norm can be approximated well by
the second algorithm. Boolean circuits expressing DNFs can be
approximated by the first one but even better by the second. In fact,
in many cases a function can be approximated by a small set of
coefficients but these coefficients do not correspond to low-order
terms. All these cases are consistent with the notes about sparse
functions in section \ref{Notes}.

\section{Notes on a theory of compositional computation}
\label{Notes}
The key property of the theory of compositional functions sketched
here is that certain deep networks can learn them avoiding the curse of
dimensionality because of the blessing of {\it compositionality} via a
small {\it effective dimension}.

We state here several comments and conjectures.

\begin{enumerate}

\item {\it General comments}
\begin{itemize}

\item  {\it Assumptions of
    the compositionality type may have more direct practical
    implications and be at least as effective as assumptions about
    function smoothness in countering the curse of dimensionality in
    learning and approximation.}

\item The estimates on the $n$--width imply that there is some
  function in either $W_m^n$ (theorem \ref{optneurtheo}) or
  $W_m^{n,2}$ (theorem \ref{deeptheo}) for which the approximation
  cannot be better than that suggested by the theorems. 


\item The main question that may be asked about the relevance of the
  theoretical results of this paper and networks used in practice has
  to do with the many ``channels'' used in the latter and with our
  assumption that each node in the networks computes a scalar function
  -- the linear combination of $r$ units (Equation \ref{ridge}). The
  following obvious but interesting extension of Theorem
  \ref{optneurtheo} to vector-valued functions says that the number of
  hidden units required for a given accuracy in each component of the
  function is the same as in the scalar case considered in our
  theorems (of course the number of weigths is larger):

\begin{corollary}
\label{vector-valued}
Let $\sigma :\R\to \R$ be infinitely differentiable, and not a
polynomial.  For a vector-valued function $f: \R^n \rightarrow \R^q$ with
components $f_i\in W_m^n, \,\, i=1,\cdots, q$ the number of hidden
units in shallow networks with $n$ inputs, $q$ outputs that provide
accuracy at least $\epsilon$ in each of the components of $f$ is \be N=
\O(\epsilon^{-n/m})\,\, .  \ee

\end{corollary}

The demonstration follows the proof of theorem \ref{optneurtheo}, see
also Appendix \ref{polapproach}. It amounts to realizing that the
hidden units (or linear combinations of them) can be equivalent to the
monomials of a generic polynomial of degree $k$ in $n$ variables that
can be used by a different set of coefficients for each of the
$f_i$. This argument of course does not mean that during learning this
is what happens; it provides one way to perform the approximation
and an associated upper bound. The corollary above leads to a simple
argument that {\it generalizes our binary tree results to standard,
  multi-channel deep convolutional networks} by introducing a set of
virtual linear units as ouputs of one layer and inputs of the next
one.
This in turn leads to the following {\it prediction}: for consistent
approximation accuracy across the layers, the rank of the weights
matrices between units in successive layers should have a rank in the
order of the number of the dimensionality in the first layer (inputs
and outputs have to be defined wrt support of the convolution kernel). This
suggests  rank-deficient weight matrices in present networks.

\item We have used polynomials (but see Appendix \ref{Booleanization})
  to prove results about complexity of approximation in the case of
  neural networks. Neural network learning with SGD may or may not
  synthesize polynomial, depending on the smoothness of the activation
  function and on the target. This is not a problem for theoretically
  establishing upper bounds on the degree of convergence because
  results using the framework on nonlinear width guarantee the
  ``polynomial'' bounds are optimal.

\item Both shallow and deep representations may or may not reflect
  invariance to group transformations of the inputs of the function
  (\cite{Soatto2011,anselmi2015theoretical}). Invariance -- also
  called weight sharing -- decreases the complexity of the network.
  Since we are interested in the comparison of shallow vs deep
  architectures, we have considered the generic case of networks (and
  functions) for which invariance is not assumed. In fact, the key
  advantage of deep vs. shallow network -- as shown by the proof of
  the theorem -- is the associated hierarchical locality (the
  constituent functions in each node are local that is have a small
  dimensionality) and {\it not invariance} (which designates shared
  weights that is nodes at the same level sharing the same
  function). One may then ask about the relation of these results with
  i-theory\cite{AnselmiPoggio2016}. The original core of i-theory
  describes how pooling can provide either shallow or deep networks
  with invariance and selectivity properties. Invariance of course
  helps but not exponentially as hierarchical locality does.

\item There are several properties that follow from the theory here which
are attractive from the point of view of neuroscience. A main one is
the robustness of the results with respect to the choice of
nonlinearities (linear rectifiers, sigmoids, Gaussians etc.) and
pooling.

\item  In a machine learning context, minimization over a
  training set of a loss function such as the square loss yields an
  empirical approximation of the regression function $p(y/x)$. Our
  hypothesis of compositionality becomes an hypothesis about the
  structure of the conditional probability function.

\end{itemize}

\item {\it Spline approximations, Boolean functions and tensors}
\begin{itemize}

\item Consider again the case of section \ref{multivariate function}
  of a multivariate function $f : [0, 1]^d \to \R$. Suppose to
  discretize it by a set of piecewise constant splines and their
  tensor products. Each coordinate is effectively replaced by $n$
  boolean variables.This results in a $d$-dimensional table with
  $N=n^d$ entries. This in turn corresponds to a boolean function
  $f: \{0,1\}^N \rightarrow \R$.  Here, the assumption of
  compositionality corresponds to compressibility of a $d$-dimensional
  table in terms of a hierarchy of $d-1$ $2$-dimensional
  tables. Instead of $n^d$ entries there are $(d-1)n^2$ entries. This
  has in turn obvious connections with HVQ (Hierarchical Vector
  Quantization), discussed in Appendix \ref{HVQ}.

\item As Appendix \ref{Booleanization} shows, every function $f$ can
  be approximated by an epsilon-close binary function $f_B$.
  Binarization of $f: \R^n \rightarrow \R$ is done by using $k$
  partitions for each variable $x_i$ and indicator functions. Thus
  $f \mapsto f_B: \{0,1\}^{kn} \rightarrow \R$ and
  $sup|f-f_B| \leq \epsilon$, with $\epsilon$ depending on $k$ and
  bounded $D f$.
\item $f_B$ can be written as a polynomial (a Walsh decomposition)
  $f_B \approx p_B$. It is always possible to associate a $p_b$ to any $f$, given $\epsilon$.

\item The binarization argument suggests a direct way to connect
  results on function approximation by neural nets with older results
  on Boolean functions. The latter are special cases of the former
  results. 

\item One can think about tensors in terms of $d$-dimensional
  tables. The framework of hierarchical decompositions of tensors --
  in particular the {\it Hierarchical Tucker format} -- is closely
  connected to our notion of compositionality. Interestingly, the
  hierarchical Tucker decomposition has been the subject of recent
  papers on Deep Learning (for instance see \cite{CohenS15}). This
  work, as well more classical papers \cite{Grasedyck}, does not
  characterize directly the class of functions for which these
  decompositions are effective.  Notice that tensor decompositions
  {\it assume} that the sum of polynomial functions of order $d$ is
  sparse (see eq. at top of page 2030 of \cite{Grasedyck}).  Our
  results provide a rigorous grounding for the tensor work related to
  deep learning. There is obviously a wealth of interesting
  connections with approximation theory that should be explored.

Notice that the notion of  {\it separation rank} of a tensor
is very closely related to the effective $r$ in Equation
\ref{Effective r}.  


\end{itemize}

\item {\it Sparsity}
\begin{itemize}
\item We suggest to define binary sparsity of $f$, in terms of the
  sparsity of the boolean function $p_B$; binary sparsity implies that
  an approximation to $f$ can be learned by non-exponential deep
  networks via binarization. Notice that if the function $f$ is
  compositional the associated Boolean functions $f_B$ is sparse; the
  converse is not true.

\item In may situations, Tikhonov regularization corresponds to {\it
    cutting high order Fourier coefficients}. {\it Sparsity} of the
  coefficients subsumes Tikhonov regularization in the case of a
  Fourier representation. Notice that as an effect the number of
  Fourier coefficients is reduced, that is
  trainable parameters, in the approximating trigonometric polynomial.  {\it Sparsity of
    Fourier coefficients} is a  general constraint for learning
  Boolean functions.

\item {\it Sparsity in a specific basis.} A set of functions may be defined to be {\it sparse in a
    specific basis} when the number of parameters necessary for its
  $\epsilon$-approximation increases less than exponentially with the
  dimensionality. An open question is the appropriate definition of
  sparsity. The notion of sparsity we suggest here is the effective
  $r$ in Equation \ref{Effective r}. For a general function
  $r \approx k^{n}$; we may define sparse functions those for which
  $r << k^{n}$ in

\begin{equation}
f(x) \approx P^*_k(x) = \sum _{i=1}^r p_i (\scal{w_i}{x}).
\end{equation}

{\it where $P^*$ is a specific polynomial that approximates $f(x)$
  within the desired $\epsilon$}. Notice that the polynomial $P^*_k$
can be a sum of monomials or a sum of, for instance, orthogonal
polynomials with a total of $r$ parameters.  In general, sparsity
depends on the basis and one needs to know the basis and the type of
sparsity to exploit it in learning, for instance with a deep network
with appropriate activation functions and architecture.

There are function classes that are sparse in every bases. Examples
are compositional functions described by a binary tree graph.

\end{itemize}

\item {\it The role of compositionality in generalization by multi-class deep networks}

Most of the succesfull neural networks exploit compositionality for
better generalization in an additional important way (see
\cite{Tenenbaumetal2016}). Suppose that the mappings to be learned in
a family of classification tasks (for instance classification of
different object classes in Imagenet) may be approximated by
compositional functions such as
$f(x_1, \cdots, x_n) = h_l \cdots (h_{21}(h_{11} (x_1, x_2),
h_{12}(x_3, x_4)), \allowbreak h_{22}(h_{13}(x_5, x_6), h_{14}(x_7,
x_8) \cdots )) \cdots) $, where $h_l$ depends on the task (for
instance to which object class) but all the other constituent
functions $h$ are common across the tasks.  Under such an assumption,
multi-task learning, that is training simultaneously for the different
tasks, forces the deep network to ``find'' {\it common constituent
  functions}. Multi-task learning has theoretical advantages that
depends on compositionality: the sample complexity of the problem can
be significantly lower (see \cite{{Maurer2015}}). The Maurer's
approach is in fact to consider the overall function as the
composition of a preprocessor function common to all task followed by
a task-specific function. As a consequence,  the {\it
  generalization error}, defined as the difference between expected
and empirical error, averaged across the $T$ tasks, is
bounded with probability at least $1-\delta$ (in
the case of finite hypothesis spaces) by
\begin{equation}
\frac{1}{\sqrt{2M}}\sqrt{ln |\mathcal{H}| +\frac{ln |\mathcal{G}| +ln (\frac{1}{\delta})}{T}},
\end{equation}
\noindent where $M$ is the size of the training set, $\mathcal{H}$ is
the hypothesis space of the common classifier and $\mathcal{G}$ is the
hypothesis space of the system of constituent functions, common across
tasks.

The improvement in generalization error because of the multitask
structure can be in the order of the square root of the number of
tasks (in the case of Imagenet with its $1000$ object classes the
generalization error may tyherefore decrease bt a factor
$\approx 30$). It is important to emphasize the dual advantage here of
compositionality, which a) reduces generalization error by decreasing
the complexity of the hypothesis space $\mathcal{G}$ of compositional
functions relative the space of non-compositional functions {\it and}
b) exploits the multi task structure, that replaces $ln |\mathcal{G}|$
with $\frac{ln |\mathcal{G}|}{T}$.

We conjecture that the good generalization exhibited by deep
convolutional networks in multi-class tasks such as CiFAR and Imagenet
are due to three factors:

\begin{itemize}
\item the regularizing effect of SGD
\item the task is compositional
\item the task is multiclass.
\end{itemize}

\item{\it Deep Networks as memories}

Notice that independently of considerations of generalization, deep
compositional networks are expected to be {\it very efficient
  memories} -- in the spirit of hierarchical vector quantization --
for associative memories reflecting compositional rules (see Appendix
\ref{HVQ} and \cite{AnselmiTan2015}). Notice that the advantage with
respect to shallow networks from the point of view of memory capacity
can be exponential (as in the example after Equation
\ref{ratiosamplesize} showing
$m_{shallow} \approx {10}^{{10}^4} m_{deep}$).

\item {\it Theory of computation, locality and compositionality}
\begin{itemize}

\item From the computer science point of view, feedforward multilayer
  networks are equivalent to finite state machines running for a
  finite number of time steps\cite{Shalev-Shwartz2014,
    poggio1980}. This result holds for almost any fixed nonlinearity
  in each layer. Feedforward networks are equivalent to cascades
  without loops (with a finite number of stages) and all other forms
  of loop free cascades (i.e. McCulloch-Pitts nets without loops,
  perceptrons, analog perceptrons, linear threshold machines). Finite
  state machines, cascades with loops, and difference equation systems
  which are Turing equivalent, are more powerful than multilayer
  architectures with a finite number of layers. The latter networks,
  however, are practically universal computers, since every machine we
  can build can be approximated as closely as we like by defining
  sufficiently many stages or a sufficiently complex single
  stage. Recurrent networks as well as
  differential equations are Turing universal.

  In other words, all computable functions (by a Turing machine) are
  recursive, that is composed of a small set of primitive
  operations. In this broad sense all computable functions are
  compositional (composed from elementary functions). Conversely a
  Turing machine can be written as a compositional function
  $y=f^{(t}(x,p)$ where $f: Z^n \times P^m \mapsto Z^h \times P^k$,
  $P$ being parameters that are inputs and outputs of $f$. If $t$ is
  bounded we have a finite state machine, otherwise a Turing machine.
  in terms of elementary functions. As mentioned above, each layer in
  a deep network correspond to one time step in a Turing machine. In a
  sense, this is sequential compositionality, as in the example of
  Figure \ref{example_3_functions} c.  The hierarchically local
  compositionality we have introduced in this paper has the flavour of
  compositionality in space.

Of course, since any function can be approximated by polynomials, and a
polynomial can always be calculated  using a recursive procedure and a
recursive procedure can always be unwound as a ``deep network'', any
function can always be approximated by a compositional function of a
few variables. However, generic compositionality of this type does not
guarantee good approximation properties by deep networks.

\item Hierarchically local compositionality  can be related to the notion of {\em
local  connectivity} of a network.  Connectivity is a key property in
  network computations. Local processing may be a key constraint also
  in neuroscience. One of the natural measures of connectivity that
  can be introduced is the {\it order} of a node defined as {\it the
    number of its distinct inputs}. The {\it order of a network is
    then the maximum order among its nodes}. The term order dates back
  to the Perceptron book (\cite{MinskyPapert}, see also
  \cite{poggio1980}). From the previous observations, it follows that
  a {\it hierarchical network of order at least $2$ can be universal.}
  In the {\it Perceptron} book many interesting visual computations
  have low order (e.g. recognition of isolated figures). The message
  is that they can be implemented in a single layer by units that have
  a small number of inputs. More complex visual computations require
  inputs from the full visual field. A hierarchical network can
  achieve effective high order at the top using units with low order.
  The network architecture of Figure \ref{example_3_functions} b) has
  low order: each node in the intermediate layers is connected to just
  2 other nodes, rather than (say) all nodes in the previous layer
  (notice that the connections in the trees of the figures may reflect
  linear combinations of the input units).

\item Low order may be a key constraint for cortex. If it captures
  what is possible in terms of connectivity between neurons, it may
  determine by itself the hierarchical architecture of cortex which in
  turn may impose compositionality to language and speech.

\item The idea of functions that are compositions of ``simpler''
  functions extends in a natural way to recurrent computations and
  recursive functions. For instance $h(f^{(t)}g((x)))$ represents $t$
  iterations of the algorithm $f$ ($h$ and $g$ match input and output
  dimensions to $f$).

\end{itemize}
\end{enumerate}
\section{Why are compositional functions so common?}

Let us provide a couple of simple examples of compositional
functions.  Addition is compositional but the degree of approximation does
not improve by decomposing addition in different layers of a network;
all linear operators are compositional with no advantage for deep
networks; multiplication as well as the AND operation (for Boolean variables) is
the prototypical compositional function that provides an advantage to
deep networks. So compositionality is not enough: we need certain
sublasses of compositional functions (such as the hierarchically local
functions we described) in order to avoid the curse of dimensionality.

It is not clear, of course,  why problems encountered in practice should match this class
of functions. Though we and others have argued that the explanation
may be in either the physics or the neuroscience of the brain, these
arguments (see Appendix \ref{Does Physics or Neuroscience imply
  compositionality?}) are not (yet) rigorous. Our conjecture at present is that
compositionality is imposed by the wiring of our cortex and is
reflected in language and the common problems we worry about. Thus
compositionality of several -- but not all -- computations on
images many reflect the way we describe and think about them.

\vspace{1in}

{\bf Acknowledgment}

This work was supported by the Center for Brains, Minds and Machines
(CBMM), funded by NSF STC award CCF – 1231216. HNM was supported in
part by ARO Grant W911NF-15-1-0385. We thank O. Shamir for useful
emails that prompted us to clarify our results in the context of lower
bounds and for pointing out a number of typos and other mistakes.

\bibliographystyle{ieeetr}
\small
\bibliography{Boolean}
\normalsize

\newpage

\setcounter {section} {0}
\begin{center}
{\Huge Appendix: observations, theorems and conjectures}
\end{center}

\section{Boolean Functions}
\label{Boolean Functions}

One of the most important tools for theoretical computer scientists
for the study of functions of $n$ Boolean variables, their related
circuit design and several associated learning problems, is the
Fourier transform over the Abelian group $\mathcal{Z}^n_2$ . This is
known as Fourier analysis over the Boolean cube $\{-1,1\}^n$. The
Fourier expansion of a Boolean function $f: \{-1,1 \}^n \to \{-1,1\}$
or even a real-valued Boolean function $f: \{-1,1 \}^n \to [-1,1]$ is
its representation as a real polynomial, which is multilinear because
of the Boolean nature of its variables. Thus for Boolean functions
their Fourier representation is identical to their polynomial
representation. In this paper we use the two terms
interchangeably. Unlike functions of real variables, the full finite
Fourier expansion is exact, instead of an approximation. There is no
need to distingush between trigonometric and real polynomials. Most of
the properties of standard harmonic analysis are otherwise preserved,
including Parseval theorem. The terms in the expansion correspond to
the various monomials; the low order ones are parity functions over
small subsets of the variables and correspond to low degrees and low
frequencies in the case of polynomial and Fourier approximations,
respectively, for functions of real variables.

\section{Does Physics or Neuroscience imply compositionality?}
\label{Does Physics or Neuroscience imply compositionality?}

It has been often argued that not only text and speech are
compositional but so are images. There are many phenomena in nature
that have descriptions along a range of rather different scales. An
extreme case consists of fractals which are infinitely self-similar,
iterated mathematical constructs. As a reminder, a self-similar object
is similar to a part of itself (i.e. the whole is similar to one or
more of the parts). Many objects in the real world are statistically
self-similar, showing the same statistical properties at many scales:
clouds, river networks, snow flakes, crystals and neurons branching.
A relevant point is that the shift-invariant scalability of image
statistics follows from the fact that objects contain smaller clusters
of similar surfaces in a selfsimilar fractal way. Ruderman
\cite{Ruderman1997} analysis shows that image statistics reflects what
has been known as the property of compositionality of objects and
parts: parts are themselves objects, that is selfsimilar clusters of
similar surfaces in the physical world.  Notice however that, from the
point of view of this paper, it is misleading to say that an image is
compositional: in our terminology a {\it function on an image may be
  compositional but not its argument}. In fact, functions to be
learned may or may not be compositional even if their input is an
image since they depend on the input but also on the task (in the
supervised case of deep learning networks all weights depend on $x$
and $y$). Conversely, a network may be given a function which can be
written in a compositional form, independently of the nature of the
input vector such as the function ``multiplication of all scalar
inputs' components''. Thus a more reasonable statement is that ``many
natural questions on images correspond to algorithms which are
compositional''. Why this is the case is an interesting open
question. An answer inspired by the condition of ``locality'' of the
constituent functions in our theorems and by the empirical success of
deep convolutional networks has attracted some attention. The starting
observation is that in the natural sciences-- physics, chemistry, biology -- many
phenomena seem to be described well by processes that {\it that take place at a
  sequence of increasing scales and are local at each scale, in the sense
  that they can be described well by neighbor-to-neighbor
  interactions}.

Notice that this is a much less stringent requirement than
renormalizable physical processes \cite{TegmarkLin2016} where the {\it
  same} Hamiltonian (apart from a scale factor) is required to
describe the process at each scale (in our observation above, the
renormalized Hamiltonian only needs to remain local at each
renormalization step)\footnote{Tegmark and Lin \cite{TegmarkLin2016} have also
  suggested that a sequence of generative processes can be regarded as
  a Markov sequence that can be inverted to provide an inference
  problem with a similar compositional structure. The resulting
  compositionality they describe does not, however, correspond to our
  notion of {\it hierarchical locality} and thus our theorems cannot
  be used to support their claims.}. As discussed previously
\cite{poggio2015December} hierarchical locality may be related to
properties of basic physics that imply local interactions {\it at each
  level in a sequence of scales}, possibly different at each level. To
complete the argument one would have then to assume that several
different questions on sets of natural images may share some of the
initial inference steps (first layers in the associated deep network)
and thus share some of features computed by intermediate layers of a
deep network. In any case, at least two open questions remain that
require formal theoretical results in order to explain the connection
between hierarchical, local functions and physics:

\begin{itemize}
\item can hierarchical locality be derived from the Hamiltonians of
  physics? In other words, under which conditions does coarse graining
  lead to local Hamiltonians? According to Tegmark locality of the
  Hamiltonian ensures locality at every stage of coarse graining.
\item is it possible to formalize how and when the local hierarchical structure of
  {\it computations on images} is related to the hierarchy of local physical
  process that describe the physical world represented in the image?
\end{itemize}

It seems to us that the above set of arguments is unsatisfactory and
unlikely to provide the answer. One of the arguments is that {\it
  iterated local functions} (from $\R^n$ to $\R^n$ with $n$ increasing
without bound) can be {\it Turing universal} and thus can simulate any
physical phenomenon (as shown by the game {\it Life} which is local
and Turing universal). Of course, this does not imply that the
simulation will be efficient but it weakens the physics-based
argument. An alternative hypothesis in fact is that locality across
levels of explanation originates from the structure of the brain -- wired, say, similarly to
convolutional deep networks -- which is then forced to use local
algorithms of the type shown in Figure \ref{ScalableOperator}. Such
local algorithms allow the organism to survive because enough of the
key problems encountered during evolution can be solved well enough by
them. So instead of claiming that all questions on the physical world
are local because of its physics we believe that local algorithms are
good enough over the distribution of evolutionary relevant problems.
From this point of view locality of algorithms follows from the need
to optimize local connections and to reuse computational elements.
Despite the high number of synapses on each neuron it would be
impossible for a complex cell to pool information across all the
simple cells needed to cover an entire image, as needed by a single
hidden layer network.

\section{Splines: some notes}

\subsection{Additive and Tensor Product Splines}\label{AppHaus}

 Additive and  tensor product splines are two alternatives to radial
  kernels for multidimensional function approximation. It is well
  known that the three techniques follow from classical Tikhonov
  regularization and correspond to one-hidden layer networks with
  either the square loss or the SVM loss.

We recall the extension of classical splines approximation techniques
to multidimensional functions. The setup is due to Jones et
al. (1995) \cite{GirJonPog95}.

\subsubsection{Tensor product splines \label{sec:tensor}}

The best-known multivariate extension of one-dimensional splines is
based on the use of radial kernels such as the Gaussian or the
multiquadric radial basis function.  An alternative to choosing a
radial function is a {\it tensor product} type of basis function, that
is a function of the form

$$
K(x) = \Pi_{j=1}^d k(x_j)
$$

\noindent
where $x_j$ is the $j$-th coordinate of the vector $x$ and
$k(x)$ is the inverse Fourier transform associated with a Tikhonov stabilizer
(see \cite{GirJonPog95}).

\noindent
We notice that the choice of the Gaussian basis function for $k(x)$
leads to a Gaussian radial approximation scheme with $K(x) = e^{- \|
x\|^2}$.

\subsubsection{Additive splines \label{subsec:additive}}

Additive approximation schemes can also be derived in the framework of
regularization theory. With additive approximation we mean an approximation of
the form

\begin{equation}
f(x) = \sum_{\mu = 1}^d f_\mu(x^\mu)
\label{eq:additive}
\end{equation}

\noindent
where $x^\mu$ is the $\mu$-th component of the input vector $x$
and the $f_\mu$ are one-dimensional functions that will be defined as
the {\it additive components} of $f$ (from now on Greek letter indices
will be used in association with components of the input vectors).
Additive models are well known in statistics (at least since Stone,
1985) and can be considered as a generalization of linear models. They
are appealing because, being essentially a superposition of
one-dimensional functions, they have a low complexity, and they share
with linear models the feature that the effects of the different
variables can be examined separately. The resulting scheme is very
similar to Projection Pursuit Regression.  We refer to
\cite{GirJonPog95} for references and discussion of how such approximations follow
from regularization.

\smallskip

Girosi et al. \cite{GirJonPog95} derive an approximation scheme of the
form (with $i$ corresponding to spline knots -- which are free
parameters found during learning as in free knots splines - and $\mu$
corresponding to new variables as linear combinations of the original
components of $x$):

\begin{equation}
f(x) = \sum_{\mu = 1}^{d^\prime} \sum_{i=1}^n c_i^\mu K(\scal{t^{\mu}}{x} - b^\mu) = \sum_{\mu = 1} \sum_{i=1} c_i^\mu K(\scal{t^{\mu}}{x} - b_i^\mu)~.
\label{eq:add-network}
\end{equation}

Note that the above can be called spline only with a stretch of the
imagination: not only the $w_\mu$ but also the $t^\mu$ depend
on the data in a very nonlinear way. In particular, the $ t^\mu$ may
not correspond at all to actual data point. The approximation could be
called ridge approximation and is related to projection pursuit.
\noindent
When the basis function $K$ is the absolute value that is $K(x-y) =
|x-y|$ the network implements piecewise linear splines.

\subsection{Hierarchical Splines}

Consider an additive approximation scheme (see subsection) in which a
function of $d$ variables is approximated by an expression such as

\begin{equation}
f( x) = \sum_{i}^d \phi_i(x^i)
\label{eq:add}
\end{equation}

\noindent
where $x^i$ is the $i$-th component of the input vector $x$ and the
$\phi_i$ are one-dimensional spline approximations. For linear
piecewise splines $\phi_i(x_i) = \sum_j c_{ij} |x^i-b^j_i|$. Obviously
such an approximation is not universal: for instance it cannot
approximate the function $f(x,y)=xy$. The classical way to deal with
the problem is to use tensor product splines. The new alternative that
we propose here is {\it hierarchical additive splines}, which in the
case of a $2$-layers hierarchy has the form

\begin{equation}
f(x) = \sum_j^K \phi_j (\sum_{i}^d \phi_i(x^i)).
\label{eq:add2}
\end{equation}

\noindent and which can be clearly extended to an arbitrary depth. The
intuition is that in this way, it is possible to obtain approximation
of a function of several variables from functions of one variable
because interaction terms such as $xy$ in a polynomial approximation
of a function $f(x,y)$ can be obtained from terms such
as $e^{log(x)+log(y)}$.

We start with a lemma about the relation between linear rectifiers,
which do not correspond to a kernel, and absolute value, which is a kernel.\\\\

\noindent {\bf Lemma 1} {\it Any given superposition of linear
  rectifiers $\sum_i c'_i (x - b^{'i})_+$ with $c'_i, b^{'i}$ given, can
  be represented over a finite interval in terms of the absolute value
  kernel with appropriate weights. Thus there exist $c_i, b^i$ such
  that $\sum_i c'_i (x - b^{'i})_+ = \sum_i c_i |x -
  b^i|$.}\\\\
\noindent
The proof follows from the facts that a) the superpositions of ramps
is a piecewiselinear function, b) piecewise linear functions can be
represented in terms of linear splines and c) the kernel corresponding
to linear splines in one dimension is the absolute value
$K(x,y)=|x-y|$.

Now consider two layers in a network in which we assume degenerate pooling for
simplicity of the argument. Because of Lemma 1, and because weights
and biases are arbitrary we assume that the the nonlinearity in each
edge is the absolute value.  Under this assumption, unit $j$ in the
first layer, before the non linearity, computes
\begin{equation}
f^{j}(x) = \sum_{i = 1} c^j_i |\scal{t^{i}}{x} - b^{i}|,
\label{eq:splines-network}
\end{equation}
where $x$ and $w$ are vectors and the $t^i$ are real numbers. Then the
second layer output can be calculated with
the nonlinearity $|\cdots|$ instead of $(\cdot)^{2}$.

In the case of a network with two inputs $x,y$ the effective output
after pooling at the first layer may be
$\phi^{(1)}(x,y)=t_1 |x+b_1|+t_2 |y+b_2|$, that is the linear
combination of two ``absolute value'' functions. At the second layer
terms like $\phi^{(2)}(x,y)=|t_1 |x+b_1|+t_2 |y+b_2|+b_3|$ may
appear. The output of a second layer still consists of hyperplanes,
since the layer is a kernel machine with an output which is always a
piecewise linear spline.

Networks implementing tensor product splines are universal in the
sense that they approximate any continuous function in an interval,
given enough units. Additive splines on linear combinations of the
input variables of the form in eq. \eqref{eq:splines-network} are also universal (use Theorem 3.1 in
\cite{Pinkus1999}). However additive splines on the individual
variables are not universal while hierarchical additive splines are:

\vspace{0.1in}

{\bf Theorem} {\it Hierarchical additive splines networks are
  universal.}\\\\
\label{TheoremUniversality}

\section{On multivariate function approximation}
\label{multivariate function}

Consider a multivariate function $f : [0, 1]^d \to \R$ discretized by
tensor basis functions:

\begin{equation}
\phi_{(i_1,...,i_d)}(x_1, . . . , x_d) := \prod_{\mu=1}^d \phi_{i_\mu} (x_\mu), 
\end{equation}

\noindent with $\phi_{i_\mu} : [0, 1] \to \R, 1 \leq i_\mu \leq n_\mu, 1 \leq \mu \leq
d$

\noindent to provide

\begin{equation}
f(x_1, . . . , x_d) = \sum_{i_1=1}^{n_1} \cdots \sum_{i_d=1}^{n_d}
c(i_1,...,i_d) \phi{(i_1,...,i_d)}(x_1, . . . , x_d).
\end{equation}

The one-dimensional basis functions could be polynomials (as above),
indicator functions, polynomials, wavelets, or  other sets of basis
functions. The total number $N$ of basis functions scales
exponentially in $d$ as $N =\prod_{\mu=1}^d n_{\mu}$ for a fixed
smoothness class $m$ (it scales as $\frac{d}{m}$).

We can regard neural networks as implementing some  form of this general
approximation scheme. The problem is that the type of operations
available in the  networks are limited. In particular, most of the
networks do not include the product operation (apart from ``sum-product''
networks also called ``algebraic circuits'') which is needed for the
straightforward implementation of the
tensor product approximation described above. Equivalent
implementations can be achieved however. In the next two sections we
describe how networks with a univariate ReLU nonlinearity may perform
multivariate function approximation with a polynomial basis and with a
spline basis respectively. The first result is known and we give it
for completeness. The second is simple but new.

\subsection{Neural Networks: polynomial viewpoint}
\label{polapproach}

One of the choices listed above leads to polynomial basis
functions. The standard approach to prove degree of approximations
uses polynomials. It can be summarized in three steps:

\begin{enumerate}
\item Let us denote with $\mathcal{H}_k$ the linear space of homogeneous
  polynomials of degree $k$ in $\R^n$ and with
  $P_k = \bigcup _{s=0}^k \mathcal{H}_s$ the linear space of
  polynomials of degree at most $k$ in $n$ variables. Set
  $r=\binom{n-1+k}{k} = dim \mathcal{H}_k$ and denote by $\pi_k$ the
  space of univariate polynomials of degree at most $k$. We recall that
  the number of monomials in a polynomial in $d$ variables with total
  degree $\leq N$ is $\binom{d+N}{d}$ and can be written as a linear
  combination of the same number of terms of the form
  $(\scal{w}{x} +b)^N$.  

We first prove that

\begin{equation}
P_k(x) = span ((\scal{w^i}{x})^s: i=1, \cdots, r,\\\ s=1, \cdots, k
\end{equation}

\noindent and thus, with, $p_i \in \pi_k$,

\begin{equation}
P_k(x) = \sum _{i=1}^r p_i (\scal{w_i}{x}).
\label{Effective r}
\end{equation}

Notice that the effective $r$, as compared with the theoretical $r$
which is of the order $r \approx k^{n}$, is closely related to the {\it separation
rank} of a tensor. Also notice that a polynomial of degree $k$ in $n$
variables can be represented exactly by a network with $r=k^n$ units.

\item Second, we prove that each univariate polynomial can be
  approximated on any finite interval from

\begin{equation}
\mathcal{N}(\sigma)= span \{\sigma(\lambda t - \theta)\}, \lambda,
\theta \in \R
\end{equation}

\noindent in an appropriate norm.

\item The last step is to use classical results about approximation
  by polynomials of  functions in a Sobolev space:
\begin{equation}
E(\mathcal{B}^m_p; P_k; L_p) \leq C k^{-m}
\label{rate}
\end{equation}

\noindent where $\mathcal{B}^m_p$ is the Sobolev space of functions
supported on the unit ball in $\R^n$.
\end{enumerate} 

The key step from the point of view of possible implementations by a
deep neural network with ReLU units is step number $2$. A univariate
polynomial can be synthesized -- in principle -- via the linear
combination of ReLUs units as follows. The limit of the linear
combination $\frac{\sigma ((a+h)x+b) - \sigma (ax+b)}{h}$ contains the
monomial $x$ (assuming the derivative of $\sigma$ is nonzero). In a
similar way one shows that the set of shifted and dilated ridge
functions has the following property. Consider for $c_i, b_i,
\lambda_i \in \R$ the space of univariate functions

\begin{equation}
\mathcal{N}_r(\sigma)=\left\{\ \sum_{i=1}^r c_i \sigma(\lambda_i x - b_i) \right\}.
\label{Pinkus3.6}
\end{equation}

\noindent The following (see Propositions 3.6 and 3.8 in \cite{Pinkus1999}) holds

\begin{proposition}{\it If $\sigma \in \mathcal{C}(\R)$ is not a
      polynomial and  $\sigma \in C^\infty$, the closure of $\mathcal{N}$ contains the linear
      space of algebraic polynomial of degree at most $r-1$}.
\end{proposition}

Since $r \approx k^{n}$ and thus $k \approx r^{1/n}$ equation \ref{rate} gives
\begin{equation}
E(\mathcal{B}^m_p; P_k; L_p) \leq C r^{-\frac{m}{n}}.
\end{equation}

\subsection{Neural Networks: splines viewpoint}

Another choice of basis functions for discretization consists of
splines. In particular, we focus for simplicity on indicator functions
on partitions of $[0,1]$, that is piecewise constant splines. Another
attractive choice are Haar basis functions. If we focus on the binary
case, section \ref{Booleanization} tells the full story that does not
need to be repeated here. We just add a note on establishing a
partition

Suppose that $a = x_1 <x_2 \cdots < x_m = b$ are given points, and set $\Delta x$ the
  maximum separation between any two points.
\begin{itemize}
\item If $f \in C[a,b]$ then for every $\epsilon > 0$ there is a $\delta > 0$
  such that if $\Delta x < \delta$, then $|f (x)-Sf (x)| < \epsilon$
  for all $x \in [a, b]$, where $Sf$ is the spline interpolant of $f$.

\item  if $f \in \C^2[a,b]$ then for all $x \in [a,b]$
$$|f(x) - Sf (x)| \leq \frac{1}{8} (\Delta x)^2 max_{a \leq z \leq b} |f''(z)|
$$
\end{itemize}

\noindent The first part of the Proposition states that piecewise linear
interpolation of a continuous function converges to the function when
the distance between the data points goes to zero. More specifically,
given a tolerance, we can make the error less than the tolerance by
choosing $\Delta x$ sufficiently small.  The second part gives an
upper bound for the error in case the function  is smooth, which in
this case means that $f$ and its first two derivatives are
continuous.

\subsection{Non-smooth ReLUs: how deep nets may work in reality}
\label{Booleanization}

Our main theorem (\ref{deeptheo}) in this paper is based on polynomial
approximation. Because of the n-width result other approaches to
approximation cannot yield better rates than polynomial
approximation. It is, however, interesting to consider other kinds of
approximation that may better capture what deep neural network with
the ReLU activation functions implement in practice as a consequence
of minimizing the empirical risk.

Our construction shows that a network with non-smooth ReLU activation
functions can approximate any continuous function with a rate similar
to our other results in this paper. A weakness of this results wrt to
the other ones in the paper is that it is valid in the $L_2$ norm but
not in the sup norm. This weakness does not matter in practice since a
discretization of real number, say, by using $64$ bits floating point
representation, will make the class of functions a finite class for
which the result is valid also in the $L_\infty$ norm. The logic of
the argument is simple:

\begin{itemize}
\item Consider the constituent functions of the binary
tree, that is functions of two variables such as $g(x_1, x_2)$. Assume
that $g$ is Lipschitz with Lipschitz constant $L$. Then for any
$\epsilon$ it is possible to set a partition of $x_1, x_2$ on the unit
square that allows piecewise constant approximation of $g$ with
accuracy at  least $\epsilon$ in the sup norm.
\item We show then that a multilayer network of ReLU units can
  compute the required partitions in the $L_2$ norm and perform piecewise constant
  approximation of $g$. 
\end{itemize}

Notice that partitions of two variables $x$ and $y$ can in principle be chosen in
advance yielding a finite set of points
$0=:x_0 < x_1 < \cdots <x_k:=1$ and an identical set
$0=:y_0 < y_1 < \cdots <y_k:=1$.  In the extreme, there may be as
little as one partition -- the binary case. In practice, the
partitions can be assumed to be set by the architecture of the network
and optimized during learning.  The simple way to choose partitions is
to choose an interval on a regular grid. The other way is an irregular
grid optimized to the local smoothness of the function. As we will
mention later this is the difference between fixed-knots splines and
free-knots splines.
  
We describe next a specific construction.

Here is how a linear combination of  ReLUs creates a
unit that is active if $x_1 \leq x \leq x_2$ and
$y_0 \leq y \leq y_1$. Since the ReLU activation $t_+$ is a basis
for piecewise linear splines, an approximation to an indicator
function (taking the value $1$ or $0$, with knots at  $x_1$,
$x_1 + \eta$, $x_2$ $x_2 +\eta$, ) for the interval between $x_1$ and
$x_2$ can be synthesized using at most $4$ units in one layer. A
similar set of units creates an approximate indicator function for the
second input $y$. A set of $3$ ReLU's can then perform a $min$
operations between the $x$ and the $y$ indicator functions, thus
creating an indicator function in two dimensions.

In greater detail, the argument is as follows:
For any $\epsilon>0$, $0\le x_0<x_1<1$, it is easy to construct an ReLU network $R_{x_0,x_1}$ with $4$ units as described above so that
$$
\|\chi_{[x_0,x_1)}-R\|_{L^2[0,1]}\le \epsilon.
$$
We define another ReLU network with two inputs and 3 units by 
\begin{eqnarray*}
\phi(x_1,x_2)&:=&(x_1)_+ -(-x_1)_+ -(x_1-x_2)_+
=\min (x_1,x_2)\\
&=& \frac{x_1+x_2}{2}+\frac{|x_1-x_2|}{2}.
\end{eqnarray*}

Then, with $I=[x_0,x_1)\times [y_0,y_1)$, we define a two layered network with $11$ units total by
$$
\Phi_I(x,y)=\phi(R_{x_0,x_1}(x), R_{y_0,y_1}(y)).
$$
Then it is not difficult to deduce that
\begin{eqnarray*}
\|\chi_I-\Phi_I\|_{L^2([0,1]^2)}^2&\!\!\!=\!\!\!&\int_0^1\!\!\int_0^1\\
&                                            & |\min(\chi_{[x_0,x_1)}(x),\chi_{[y_0,y_1)}(y))-\\
&                                            &\min(R_{x_0,x_1}(x), R_{y_0,y_1}(y))|^2dxdy \le c\epsilon^2.
\end{eqnarray*}

Notice that in this case dimensionality is $n=2$; notice that in
general the number of units is proportional to $k^n$ which is of the
same order as $\binom{n+k}{k}$ which is the number of parameters in a
polynomial in $n$ variables of degree $k$.  The layers we described
compute the entries in the 2D table corresponding to the bivariate
function $g$. One node in the graph (there are $n-1$ nodes in a binary
tree with $n$ inputs) contains $O(k^2)$ units; the total number of
units in the network is $(n-1) O(k^2)$. This construction leads to
the following results.

\begin{proposition}
  \label{deepreleu} 
  Compositional functions on the unit cube with an associated binary
  tree graph structure and constituent functions that are Lipschitz
  can be approximated by a deep network of ReLU units within accuracy
  $\epsilon$ in the $L_2$ norm with a number of units in the order of
  $O((n-1) L \epsilon^{-2}) $, where
  $L$ is the worse -- that is the max -- of the Lipschitz constant
  among the constituent functions.
\end{proposition}

Of course, in the case of machine numbers -- the integers -- we can
think of zero as a very small positive number. In this case, the
symmetric difference ratio
$((x+\epsilon)_+ - (x-\epsilon)_+)/(2\epsilon)$ is the hard threshold
sigmoidal function if $\epsilon$ is less than this smallest positive
number. So, we have the indicator function exactly as long as we stay
away from $0$. From here, one can construct a deep network as usual.

Notice that the number of partitions in each of two variables that are
input to each node in the graph is $k=\frac{L}{\epsilon}$ where $L$ is
the Lipschitz constant associated with the function $g$ approximated
by the node. Here the role of  smoothness
is clear: the smaller $L$ is, the smaller is the number of
variables in the approximating Boolean function.  Notice that if $g
\in W^2_1$, that is $g$ has bounded first derivatives, then $g$ is
Lipschitz.  However, {\it higher order smoothness} beyond the bound on the first
derivative {\it cannot be exploited by the network} because of the
non-smooth activation function\footnote{
In the case of univariate approximation on the interval $[-1,1]$,
piecewise linear functions with inter-knot spacing $h$ gives an accuracy
of $(h^2/2)M$, where $M$ is the max absolute value of $f''$. So, a higher
derivative does lead to better approximation : we need
$\sqrt{2M/\epsilon}$ units to give an approximation of $\epsilon$. This is
a saturation though. Even higher smoothness does not help.}.

We {\it conjecture that the construction above that performs piecewise
  constant approximation is qualitatively similar to what deep
  networks may represent after training}. Notice that the partitions
we used correspond to a uniform grid set a priori depending on global
properties of the function such as a Lipschitz bound. In supervised
training of deep network the location of each partition is likely to
be optimized in a greedy way as a function of the performance on the
training set and therefore as a function of inputs and output. Our
Theorem \ref{deeptheo} and Theorem \ref{deepreleu} are obtained under
this assumption. In any case, their proofs suggest two different ways
of how deep networks could perform function approximations, the first
by using derivatives and the second by using piecewise linear
splines. In the latter case, optimizing the partition as a function of
the input-output examples correspond to {\it free-knots splines}. The
case in which the partitions depend on the inputs but not the target,
correspond to classical {\it fixed-knots splines}. As an aside, one
expects that networks with smooth ReLUs will perform  better than
networks with non-smooth ReLUs in the approximation of very smooth
functions.

Still another way to create with ReLUs a discrete table correpsonding
to the multiplication of two variables, each taking discrete values on
a bounded interval is hash the two dimensional table into a one
dimensional table. For instance assume that $x$ and $y$ take integer
values between $[0,9]$. Set the variable $z= 10 x +y$. This is a one
dimensional table equivalent to the 2-dimensional table $x \times
y$. This Cantor-like mapping idea works only if restricted to machine
numbers, that is to the integers.

\section{Vector Quantization and Hierarchical Vector Quantization}
\label{HVQ}
Let us start with the observation that a network of
radial Gaussian-like units become in the limit of $\sigma \to 0$ a
look-up table with entries corresponding to the centers. The network
can be described in terms of {\it soft Vector Quantization} (VQ) (see
section 6.3 in Poggio and Girosi, \cite{Poggio89atheory}). Notice that
hierarchical VQ (dubbed HVQ) can be even more efficient than VQ in
terms of storage requirements (see e.g. \cite{HVQ}). This suggests
that a hierarchy of HBF layers may be similar (depending on which
weights are determined by learning) to HVQ. Note that {\it compression
  is achieved when parts can be reused in higher level layers as in
  convolutional networks}. Notice that the center of one unit at level
$n$ of a ``convolutional'' hierarchy is a combinations of parts
provided by each of the lower units feeding in it. This may even
happen without convolution and pooling as shown
in the following extreme example.\\\\
\noindent
{\bf Example}
\noindent
Consider the case of kernels that are in the limit delta-like
functions (such as Gaussian with very small variance). Suppose that
there are four possible quantizations of the input $x$:
$x_1, x_2, x_3, x_4$. One hidden layer would consist of four units
$\delta(x-x_i), i=1,\cdots,4$. But suppose that the vectors
$x_1, x_2, x_3,x_4$ can be decomposed in terms of two smaller parts or
features $x'$ and $x"$, e.g.  $x_1 =x'\oplus x"$, $x_2=x'\oplus x'$,
$x_3=x"\oplus x"$ and $x_4=x"\oplus x'$. Then a two layer network
could have two types of units in the first layer $\delta(x-x')$ and
$\delta(x-x")$; in the second layer four units will detect the
conjunctions of $x'$ and $x"$ corresponding to $x_1, x_2,
x_3,x_4$. The memory requirements will go from $4N$ to $2N/2+8$ where
$N$ is the length of the quantized vectors; the latter is much smaller
for large $N$. Memory compression for HVQ vs VQ -- that is for
multilayer networks vs one-layer networks -- increases with the number
of (reusable) parts.  Thus for problems that are {\it compositional},
such as text and images, hierarchical architectures of HBF modules
minimize memory requirements.

Classical theorems (see refrences in \cite{GirPog-Kol89,GirPog-best90}
show that one hidden layer networks can approximate arbitrarily well
rather general classes of functions. A possible advantage of
multilayer vs one-layer networks that emerges from the analysis of
this paper is memory efficiency which can be critical for large data
sets and is related to generalization rates.

\section{Approximating compositional functions with shallow and deep networks: numerical experiments}

\begin{figure*}
\centering
\includegraphics[width=0.9\textwidth]{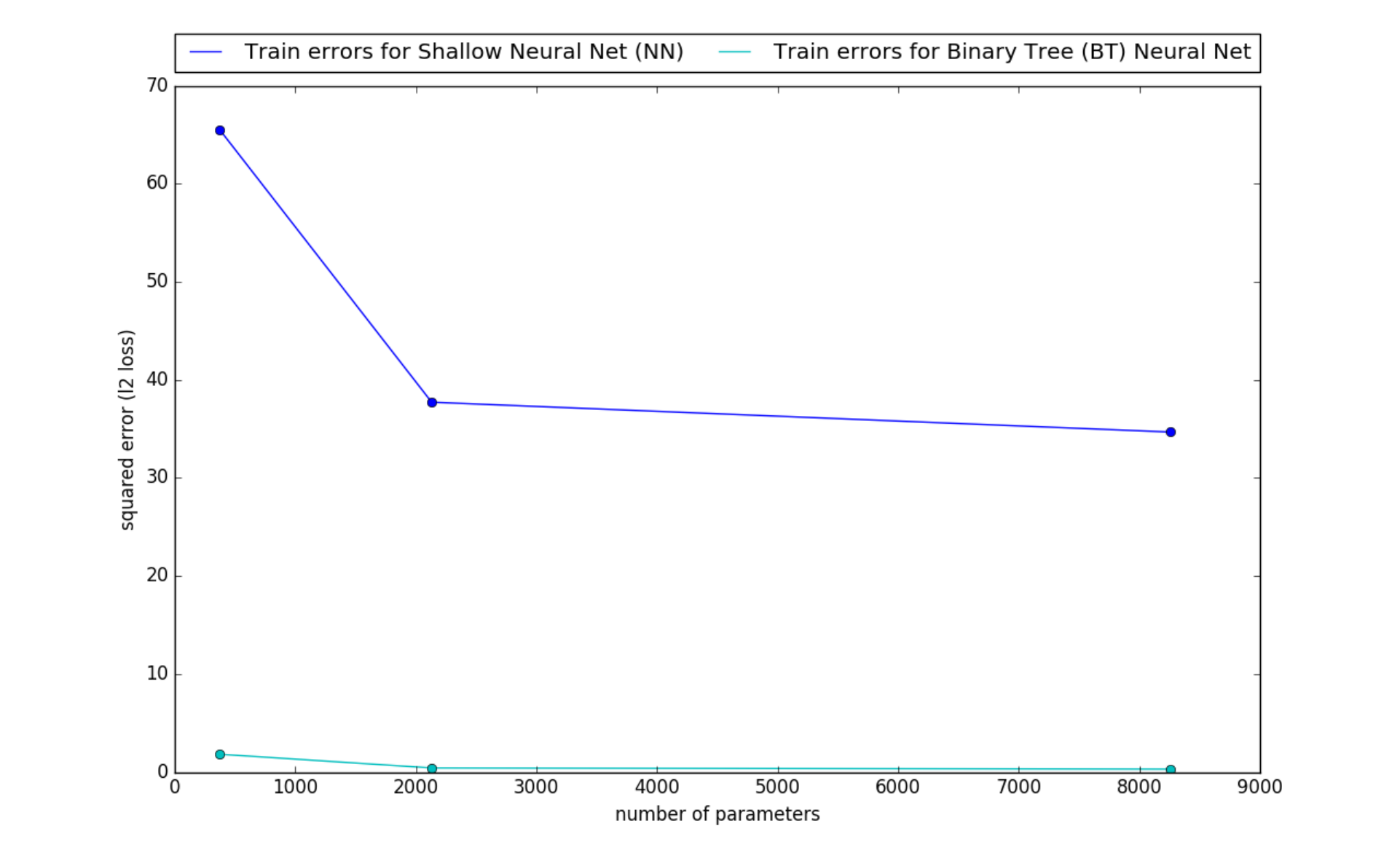}
\caption{An empirical comparison of shallow vs 3-layers binary tree
  networks in the learning of compositional functions. 
  The loss function is the the standard mean square error (MSE). There are
  several units per node of the tree. In our setup here the network
  with an associated binary tree graph was set up so that each layer
  had the same number of units.  The number of units for the shallow
  and binary tree neural network were chosen such that both
  architectures had the approximately same number of parameters. 
  The compositional function is
  $f(x_1, \cdots, x_8) = h_3(h_{21}(h_{11} (x_1, x_2), h_{12}(x_3,
  x_4)), \allowbreak h_{22}(h_{13}(x_5, x_6), h_{14}(x_7, x_8))) $
  and is approximated by a network with ReLU activations.  
  A description of the compositional function is as follows:
  the first layer units $h_{11}, h_{12}, h_{13}, h_{14}$ are equal to
  $h_1(x,y) = 0.59 cos( 1.5 \pi (x + y) )$,
  the second layer units $h_{21}, h_{22}$ are equal to $h_2(x,y) = 1.1(x + y)^2 - 1$
  and the final layer unit $h_3$ is also $h_3(x,y) = 1.1(x + y)^2 - 1$.
  The training and test sets both had 60K training examples.
  The variant of SGD that was used was the Adam \cite{DBLP:journals/corr/KingmaB14} optimizer for both experiments.  
  In order to get the best solution possible we ran 200 independent hyper
  parameter searches using random search \cite{randomhp} and then
  reported the one with the lowest training error for both tasks.  The
  hyper parameters included the step size, the decay rate, frequency
  of decay and the mini-batch size.  The exponential decay hyper
  parameters for Adam were kept fixed to the recommended values
  according to the original paper \cite{DBLP:journals/corr/KingmaB14}.  The implementations
  were based on TensorFlow \cite{tensorflow2015-whitepaper}.}
\label{Simulation_8D_cos_poly_poly}
\end{figure*}

\begin{figure*}
\centering
\includegraphics[width=0.9\textwidth]{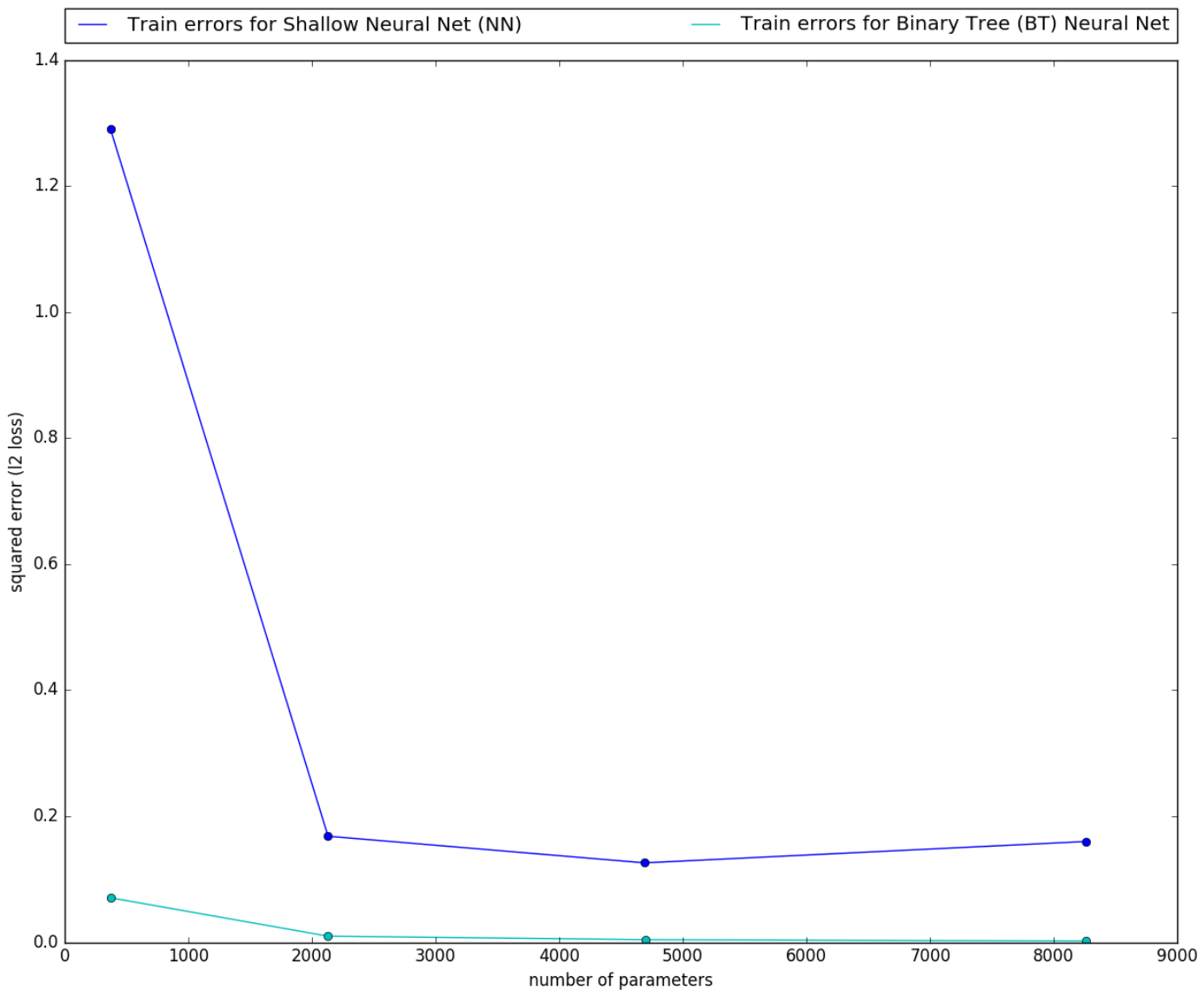}
\caption{An empirical comparison of shallow vs 3-layers binary tree
  networks in the learning of compositional functions. 
  The loss function is the the standard mean square error (MSE). There are
  several units per node of the tree. In our setup here the network
  with an associated binary tree graph was set up so that each layer
  had the same number of units.  The number of units for the shallow
  and binary tree neural network were chosen such that both
  architectures had the approximately the same number of parameters. 
  The compositional function is
  $f(x_1, \cdots, x_8) = h_3(h_{21}(h_{11} (x_1, x_2), h_{12}(x_3,
  x_4)), \allowbreak h_{22}(h_{13}(x_5, x_6), h_{14}(x_7, x_8))) $
  and is approximated by a network with ReLU activations.  
  A description of the compositional function is as follows:
  the first layer units $h_{11}, h_{12}, h_{13}, h_{14}$ are equal to
  $h_1(x,y) = 0.7 (x + 2y)^2$,
  the second layer units $h_{21}, h_{22}$ are equal to $h_2(x,y) = 0.6 (1.1x + 1.9y)^3$
  and the final layer unit $h_3$ is also $h_3(x,y) = 1.3 \sqrt{ 1.2x+1.3 }$.
  The experiment setup (training and evaluation) was the same as in \ref{Simulation_8D_cos_poly_poly}. }
\label{Simulation_8D_quad_cubic_sqrt}
\end{figure*}

\begin{figure*}
\centering
\includegraphics[width=0.9\textwidth]{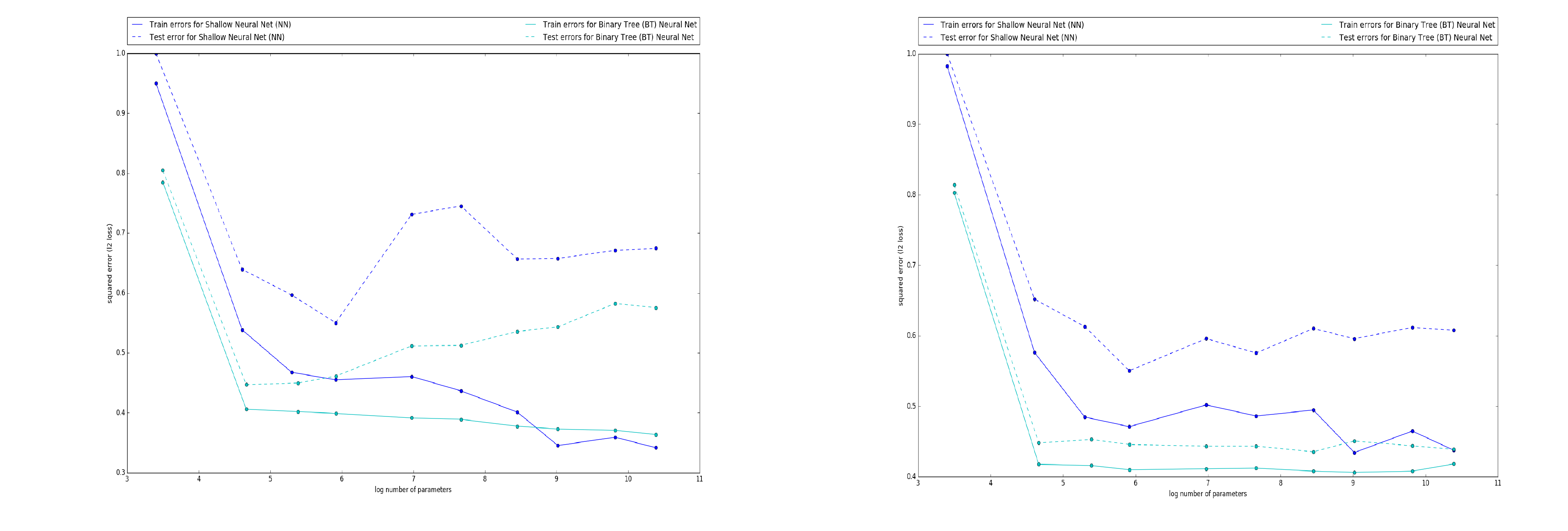}
\caption{An empirical comparison of shallow vs 3-layers binary tree
  networks in the learning of compositional functions with Gaussian noise. 
  The loss function is the the standard mean square error (MSE). There are
  several units per node of the tree. In our setup here the network
  with an associated binary tree graph was set up so that each layer
  had the same number of units.  The number of units for the shallow
  and binary tree neural network were chosen such that both
  architectures had the same number of parameters. 
  The noisy compositional function is
  $f(x_1, \cdots, x_8) = h_3(h_{21}(h_{11} (x_1, x_2), h_{12}(x_3,
  x_4)), \allowbreak h_{22}(h_{13}(x_5, x_6), h_{14}(x_7, x_8))) + \epsilon$
  and is learned by a network with
  ReLU activations. The functions $h_1$, $h_2$, $h_3$ are as described
  in Figure \ref{Simulation_8D_quad_cubic_sqrt}.
  The training and test sets both had 2K training examples.
  The variant of SGD that was used was the Adam \cite{DBLP:journals/corr/KingmaB14} optimizer for both experiments.  
  In order to get the best solution possible we ran 200 independent hyper
  parameter searches using random search \cite{randomhp}.
  For the left figure we choose the hyper parameter selection that resulted in the lowest training error.
  For the right figure choose ch the hyper parameter selection that resulted in the lowest validation error.
 For both we reported the corresponding training and test error.
  The hyper parameters included the step size, the decay rate, frequency
  of decay and the mini-batch size.  The exponential decay hyper
  parameters for Adam were the recommended values
  according to the original paper \cite{DBLP:journals/corr/KingmaB14}. 
  Its interesting to note that when the training error was used as a selection criterion,
  overfitting seemed to happen for both the shallow and binary tree network.
  However, when the validation set is used as the selection criterion overfitting is not observed anymore for either model.
  In addition, the test error for the binary tree neural network noticeably decreases while it does not for the shallow network.}
\label{fig_8D_test_train}
\end{figure*}

\begin{figure*}
\centering
\includegraphics[width=0.9\textwidth]{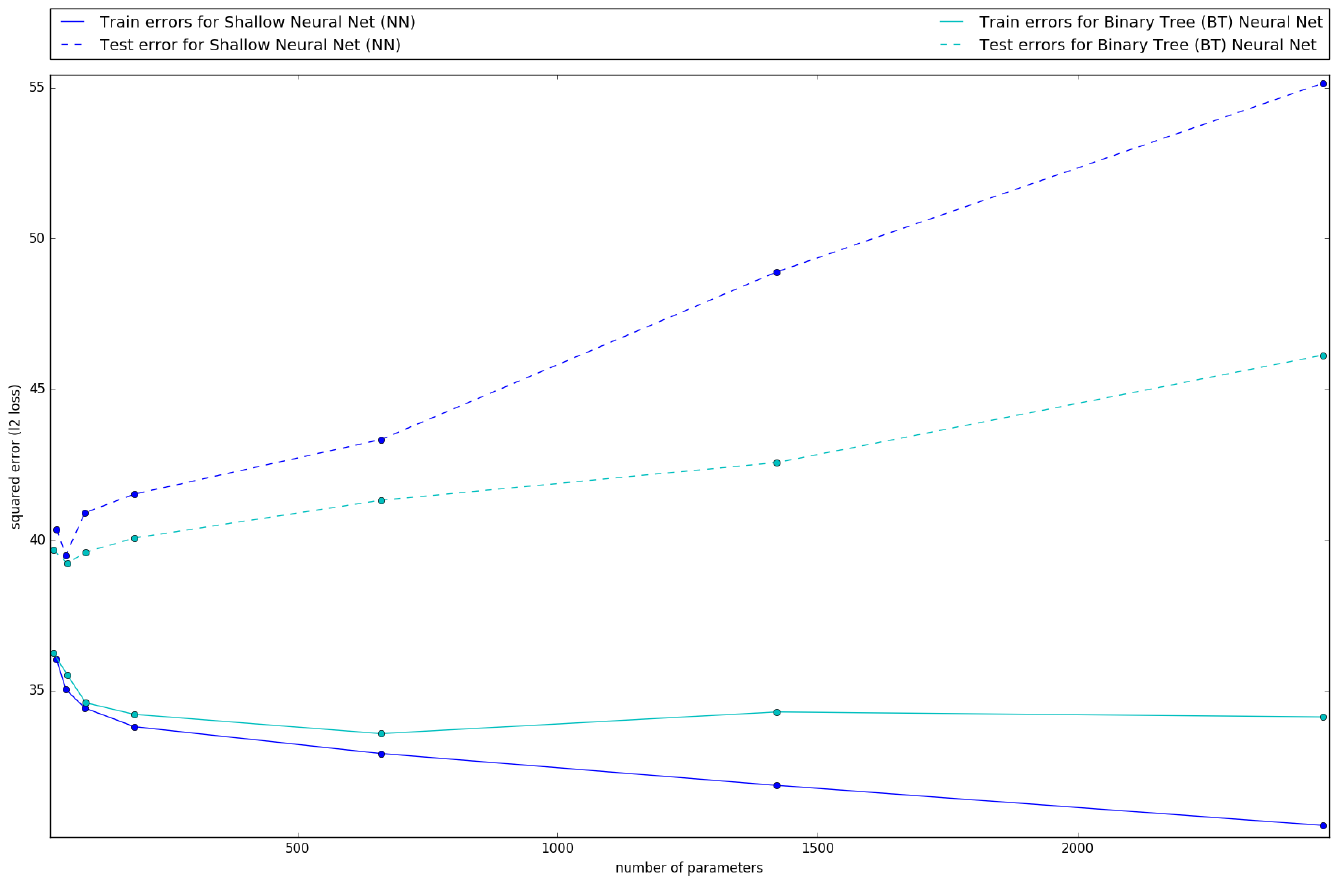}
\caption{An empirical comparison of shallow vs 2-layers binary tree
  networks in the learning of compositional functions with Gaussian noise. 
  The loss function is the the standard mean square error (MSE). There are
  several units per node of the tree.  The network
  with an associated binary tree graph was set up so that each layer
  had the same number of units.  By choosing appropriately the number of units for the shallow
  and binary tree neural networks, both
  architectures had the same number of parameters. 
  The noisy compositional function is
  $f(x_1, x_2, x_3, x_4) = h_2( h_{11}(x_1, x_2) , h_{12}(x_3, x_4) ) + \epsilon$
  and is learned by a network with
  ReLU activations. The functions $h_{11}$,$h_{12}$, $h_2$ are as described
  in Figure \ref{Functions_Simulations}.  
 The training and test sets both had 2K training examples.
  The variant of SGD that was used was the Adam \cite{DBLP:journals/corr/KingmaB14} optimizer for both experiments.  
  In order to get the best solution possible we ran 200 independent hyper
  parameter searches using random search \cite{randomhp} and then
  reported the one with the lowest training error for both tasks.  The
  hyper parameters included the step size, the decay rate, frequency
  of decay and the mini-batch size.  The exponential decay hyper
  parameters for Adam were  the recommended values
  according to the original paper \cite{DBLP:journals/corr/KingmaB14}.  The implementations
  were based on TensorFlow \cite{tensorflow2015-whitepaper}.}
\label{Simulations_noise}
\end{figure*}

\begin{figure*}
\centering
\includegraphics[width=0.9\textwidth]{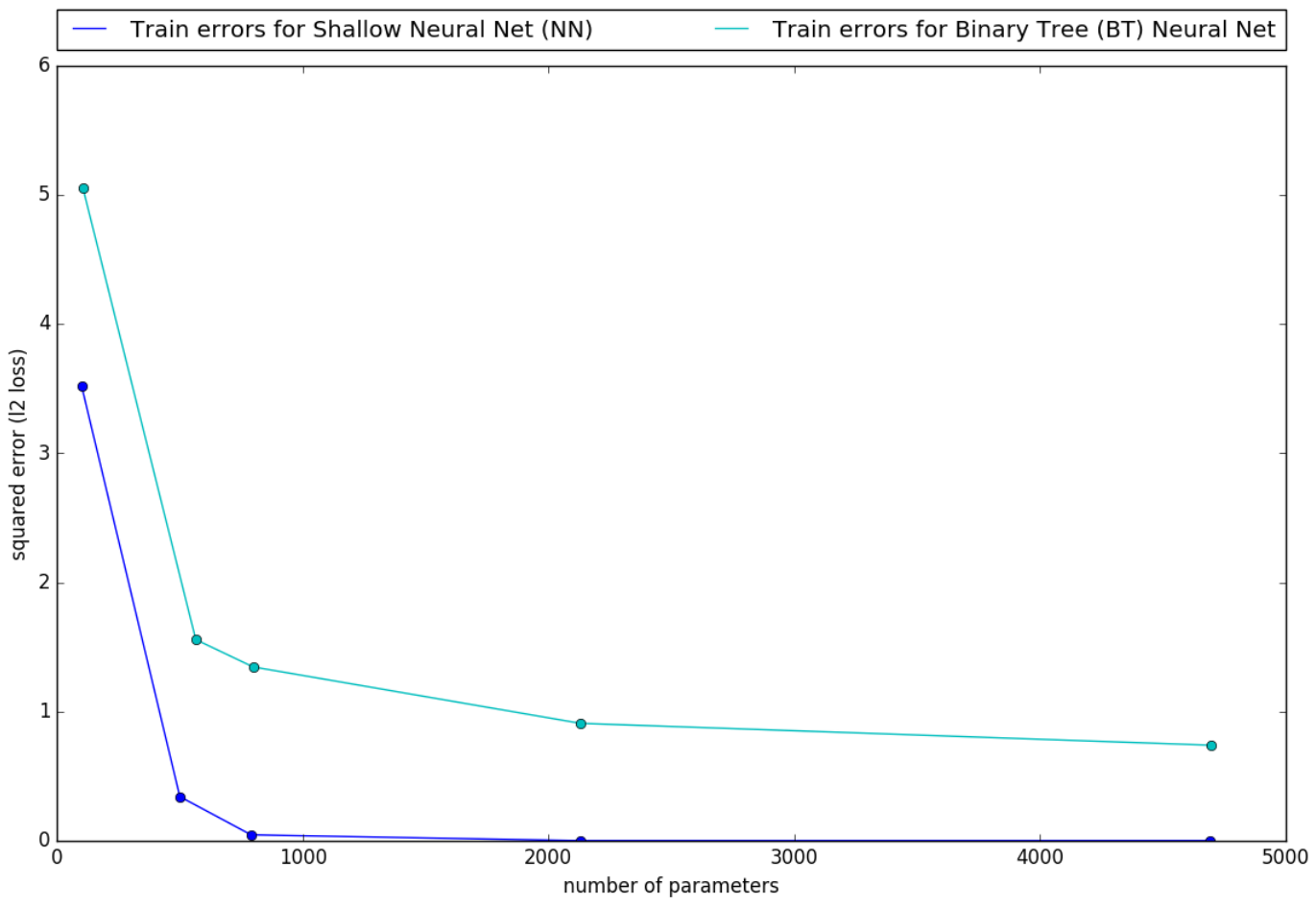}
\caption{Here we empirically show that depth of the network and
  compositionality of the function are important.
  We compare shallow vs 3-layers binary tree networks 
  in the task of approximating a function that is not compositional.
  In particular the function is a linear combination of 100 ReLU units $f(x) = \sum^{100}_{i=1} c_i (\scal{w_i}{x} + b_i)_+$.
  The loss function is the the standard mean square error (MSE).
  The training set up was the same as in figure \ref{Simulation_8D_quad_cubic_sqrt}.
  This experiment shows that when compositionality 
  is not present in the function to be approximated 
  then having depth  in the network can actually worsen performance of
  the network (confront with in \ref{Simulation_8D_quad_cubic_sqrt}).
}
\label{shallow_synthetic_100_units}
\end{figure*}

\begin{figure*}
\centering
\includegraphics[width=0.9\textwidth]{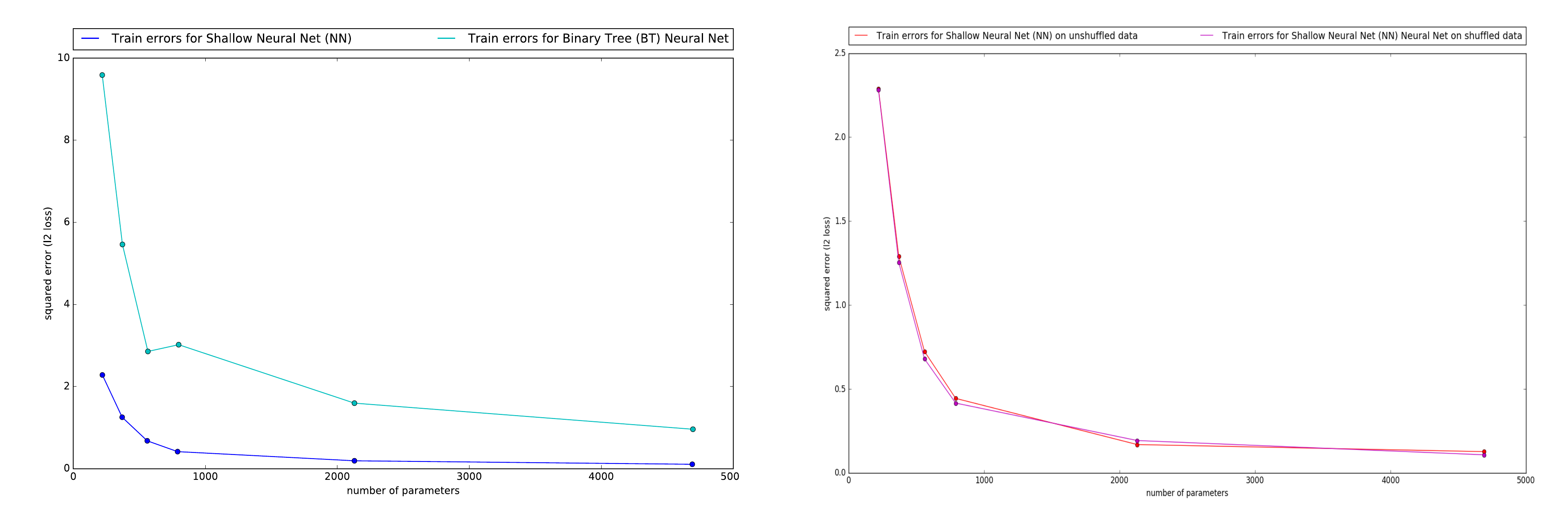}
\caption{The figure suggests again that  that depth of the network and
  compositionality of the function are important. The figure  compares shallow vs a 3-layers binary tree networks 
  in the task of approximating a function that lacks the structure of local hierarchy and compositionality.
  In particular if the compositionality of the function presented in figure \ref{Simulation_8D_quad_cubic_sqrt}
  is missing, 3-layered binary tree networks show a decreased
  approximation accuracy (see left figure).
  However, the performance of shallow networks remain completely unchanged as shown on the right figure.
 Compositionality of the function in \ref{Simulation_8D_quad_cubic_sqrt} was broken
 by shuffling the input coordinates $x \in \mathbb R^8$ randomly but 
  uniformly across all the training examples.
  More precisely a fixed shuffling function $shuffle(x)$ was selected and 
  the deep and shallow networks were trained with the data set $X = \{ (shuffle(x_i), f(x_i) \}^{60,000}_{i=1}$.
  The figure on the left shows that  the binary tree network was not able to 
  do better than the shallow network once compositionality was missing from the data.
  The figure on the right shows that this 
  did not change the performance of the shallow network.
}
\label{shuffled_vs_unshuffled}
\end{figure*}

Figures \ref{Functions_Simulations},  
\ref{Simulations} ,
\ref{Simulations_noise},
\ref{Simulations_relu2units},
\ref{Simulation_8D_cos_poly_poly},
\ref{Simulation_8D_quad_cubic_sqrt},
\ref{fig_8D_test_train}
\ref{shallow_synthetic_100_units},
and
\ref{shuffled_vs_unshuffled}
show some of our numerical experiments. 

\begin{figure*}
\centering
\includegraphics[width=0.9\textwidth]{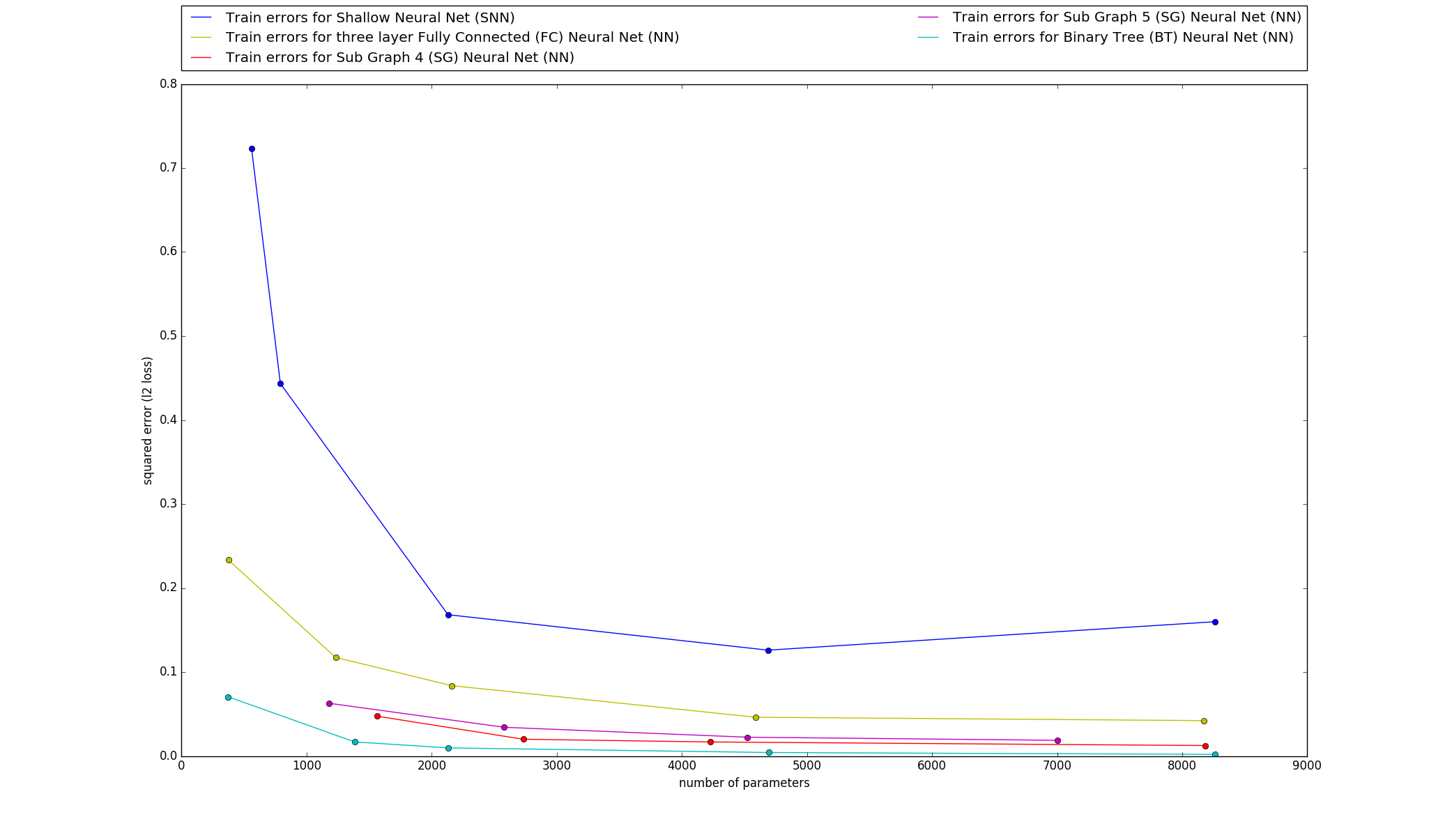}
\caption{ This figure shows a situation in which the structure of the
  neural network does not match exactly the structure of the function
  to be learned. In particular, we compare the approximation accuracy
  of networks that contain the original function as a function of
  increasing in similarity between the network graph and the function
  graph.  The function being approximated is the same as in
  \ref{Simulation_8D_quad_cubic_sqrt}.d.  The network subgraphs 4 and
  5 have the same convolution structure on the first layer with filter
  size 4.  For the second layer subgraph 5 has more connections than
  subgraph 4, which reflects the higher error of subgraph 5.  More
  precisely, subgraph 4 has a filter of size of two units with stride
  of two units while subgraph 5 has filter size of three units and
  stride of size one. This figure shows that even when the
  structure of the networks does not match exactly the function graph,
  the accuracy is  still good.  }
\label{}
\end{figure*}

\section{Compositional functions and scalable algorithms}

We formalize the requirements on the algorithms of local
compositionality is to define {\it scalable computations} as a
subclass of nonlinear discrete operators, mapping vectors from $\R^n$
into $\R^d$ (for simplicity we put in the following $d=1$). Informally
we call an algorithm $K_n: \R^n \mapsto \R$ {\it scalable} if it
maintains the same ``form'' when the input vectors increase in
dimensionality; that is, the same kind of computation takes place when
the size of the input vector changes. This motivates the following
construction. Consider a ``layer'' operator $H_{2m}: \R^{2m} \mapsto
\R^{2m-2}$ for $m \ge 1$ with a special structure that we call ``shift
invariance''.

\begin{definition}\label{scalabledef}
  For integer $m \ge 2$, an operator $H_{2m}$ is shift-invariant if
  $H_{2m} = H_m' \oplus H_m''$ where $\R^{2m}=\R^m \oplus \R^m$,
  $H'=H''$ and $H':\R^m \mapsto \R^{m-1}$.  An operator $K_{2M}
  :\R^{2M}\to \R$ is called scalable and shift invariant if
  $K_{2M}=H_2\circ \cdots H_{2M}$ where each $H_{2k}$, $1\le k\le M$,
  is shift invariant.
\end{definition}

 We observe that \textit{ scalable shift-invariant operators $K: \R^{2m}
  \mapsto \R$ have the structure $K = H_2 \circ H_4 \circ H_6 \cdots
  \circ H_{2m}$, with $H_4=H'_2 \oplus H'_2$, $H_6=H''_2 \oplus H''_2 \oplus
  H_2''$, etc.}.

\begin{figure}\centering
\includegraphics[width=0.5\textwidth]{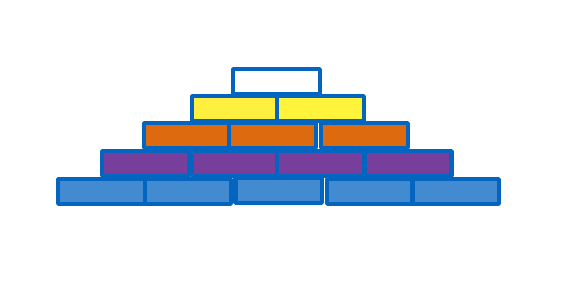}
\caption{A shift-invariant, scalable operator. Processing is from the
  bottom (input) to the top (output). Each layer consists of identical
  blocks, each block has two inputs and one output; each block is an
  operator $H_2: \R^2 \mapsto R$. The step of combining two inputs to
  a block into one output corresponds to coarse-graining of a Ising
  model.}
\label{ScalableOperator} 
\end{figure}

 Thus the structure of a {\it shift-invariant, scalable
  operator} consists of several layers; each layer consists of
identical blocks; each block is an operator $H: \R^2 \mapsto R$: see
Figure \ref{ScalableOperator} . We
note also that an alternative weaker constraint on $H_{2m}$ in
Definition~\ref{scalabledef}, instead of shift invariance, is mirror
symmetry, that is $H''= R \circ H'$, where $R$ is a
reflection. Obviously, shift-invariant scalable operator are
equivalent to shift-invariant compositional functions. Obviously the
definition can be changed in several of its details. For instance
for two-dimensional images the blocks could be operators $H: \R^5
\rightarrow \R$ mapping a neighborhood around each pixel into a real number.

The final step in the argument uses the results of previous sections
to claim that a nonlinear node with two inputs and enough units can
approximate arbitrarily well each of the $H_2$ blocks. This leads to
conclude that deep convolutional neural networks are natural
approximators of {\it scalable, shift-invariant operators}.

\section{Old-fashioned generalization bounds}

Our estimate of the number of units and parameters needed for a deep
network to approximate compositional functions with an error
$\epsilon_G$ allow the use of one of several available bounds for the
generalization error of the network to derive sample complexity
bounds. It is important to notice however that these bounds {\it do
  not} apply to the networks used today, since they assume a number of
parameters smaller than the size of the training set. We report them
for the interest of the curious reader.
Consider theorem 16.2 in \cite{AntBartlett2002}
which provides the following sample bound for a generalization error
$\epsilon_G$ with probability at least $1-\delta$ in a network in
which the $W$ parameters (weights and biases) which are supposed to
minimize the empirical error (the theorem is stated in the standard
ERM setup) are expressed in terms of $k$ bits:

\begin{equation}\label{eq:gen_bound}
M(\epsilon_G, \delta) \leq \frac{2}{\epsilon_G^2} (k W \log 2 + \log (\frac{2}{\delta}))
\end{equation}

This suggests the following comparison between shallow and deep
compositional (here binary tree-like networks). Assume a network size
that ensure the same approximation error $\epsilon$ . 

Then in order to achieve the same generalization error $\epsilon_G$,
the sample size $M_{shallow}$ of the shallow network must be much
larger than the sample size $M_{deep}$ of the deep network:

\begin{equation}
\frac{M_{deep}}{ M_{shallow}} \approx \epsilon^n.
\label{ratiosamplesize}
\end{equation}

This implies that for largish $n$ there is a (large) range of training
set sizes between $M_{deep}$ and $M_{shallow}$ for which deep networks
will not overfit (corresponding to small $\epsilon_G)$ but shallow
networks will (for dimensionality $n \approx 10^4$ and
$\epsilon \approx 0.1$ Equation \ref{ratiosamplesize} yields
$m_{shallow} \approx {10}^{{10}^4} m_{deep}$).

A similar comparison is derived if one considers the best possible
expected error obtained by a deep and a shallow network. Such an error
is obtained finding the architecture with the best trade-off between
the approximation and the estimation error. The latter is essentially
of the same order as the generalization bound implied by
inequality~\eqref{eq:gen_bound}, and is essentially the same for deep
and shallow networks, that is
\begin{equation}
\frac{rn}{\sqrt{M}},
\label{sqrt{M}}
\end{equation}
where we denoted by $M$ the number of samples.  For shallow networks,
the number of parameters corresponds to $r$ units of $n$ dimensional
vectors (plus off-sets), whereas for deep compositional networks the
number of parameters corresponds to $r$ units of $2$ dimensional
vectors (plus off-sets) in each of the $n-1$ units. Using our previous
results on degree of approximation, the number of
units giving the best approximation/estimation trade-off is
\begin{equation}
r_{shallow}\approx \left (\frac{\sqrt{M}}{n}\right)^{\frac{n}{m+n}}\quad\quad  \text{and}\quad\quad  r_{deep}\approx \left(\sqrt{M}\right)^{\frac{2}{m+2}}
\label{tradeoff}
\end{equation}
for shallow and deep networks, respectively. 
The corresponding (excess) expected errors $\mathcal E$ are 
\begin{equation}
\mathcal E _{shallow} \approx \left (
  \frac{n}{\sqrt{M}}\right)^{\frac{m}{m+n}}
\label{mess1}
\end{equation}
for shallow networks and
\begin{equation}
\mathcal E _{deep} \approx  \left
  (\frac{1}{\sqrt{M}}\right)^{\frac{m}{m+2}}
\label{mess2}
\end{equation}
for deep networks. For the expected error, as for the generalization
error, deep networks appear to achieve an exponential gain.  The above
observations hold under the assumption that the optimization process
during training finds the optimum parameters values for both deep and
shallow networks. Taking into account optimization, e.g. by stochastic
gradient descent, requires considering a further error term, but we
expect that the overall conclusions about generalization properties
for deep vs. shallow networks should still hold true.

\section{Future work}

As we mentioned in the introduction, a theory of Deep Learning should
consist of at least three main parts: Approximation, Optimization of
the Empirical Risk and Generalization. 

This paper is mainly concerned with the first part -- Approximation
Theory. We summarize briefly ongoing work, soon to be published, on
the two other parts.

{\it Theory II: Landscape of the Empirical Risk}

We assume a deep networks of the binary tree type with weight
sharing. We also assume overparametrization, that is more parameters
than data points, since this is how successful deep networks have been
used. 

In general, under these conditions, we expect that zeros of the
empirical error yield a set of quasi-polynomial equations (at the
zeros) that have an infinite number of solutions (for the network
weights) and are sparse because of ReLUs (the arguments of some of the
ReLUs are negative and thus the value of the corresponding ReLUs is zero). The
equations have the form $c_1 ABCA + c_2 ABCd c_3 A'BC'b+...=y$ with
coefficients representing components of the data vectors (one vector
per data point); the unknowns are $A,B,C,B', \cdots$.  The system of equations is linear in the
monomials with nonlinear constraints e.g. $ABCD=z_1$, $ABCd=z_2$  on
the unknown weights $A,B,C \cdots$. It is  underdetermined (more
unknowns than equations, e.g. data points) because of the assumed
overparametrization.  We use  Bezout theorem to estimate an upper
bound in the number of real zeros.





{\it Theory III: Generalization by SGD}

Part III deals with the puzzle that solutions found by ``repeat SGD''
in underdetermined situations (fewer data points e.g. equations than
unknown parameters) empirically generalize well on new data. Our key
result is to show that repeat SGD is $CV_{loo}$ stable on a empirical
risk which has many degenerate global minima. It therefore
generalizes. In other words we have the following claim

{\it Claim:
\label{thm:CV}
Suppose that $I_{S_n}[f]$ has $M$ (convex, possibly degenerate) global minima.
Then {\it repeat SGD} on $S_n$ has $CV_{loo}$, $E_{loo}$, and $EE_{loo}$ stability and generalizes.
}

Noteworthy features of generalization by ``repeat SGD'' is that
the intrinsic noise associated with SGD wrt GD is the key to its
generalization properties. The intuition is that because of the implicit
noise, SGD finds with high probability structurally stable zeros (that
are robust wrt to perturbations of the weights) and these coincide
with $CV_{loo}$stable solutions.

\end{document}